\documentclass{article}

\usepackage[eandd, preprint]{neurips_2026}
\usepackage{xcolor}
\usepackage{multirow}
\usepackage{graphicx}
\usepackage{booktabs}
\usepackage{minitoc}

\definecolor{datasetpurple}{HTML}{B80E85}
\definecolor{darkviolet}{HTML}{4B0082}   

\newcommand{\hflogo}{\raisebox{-0.2em}{\includegraphics[height=1em]{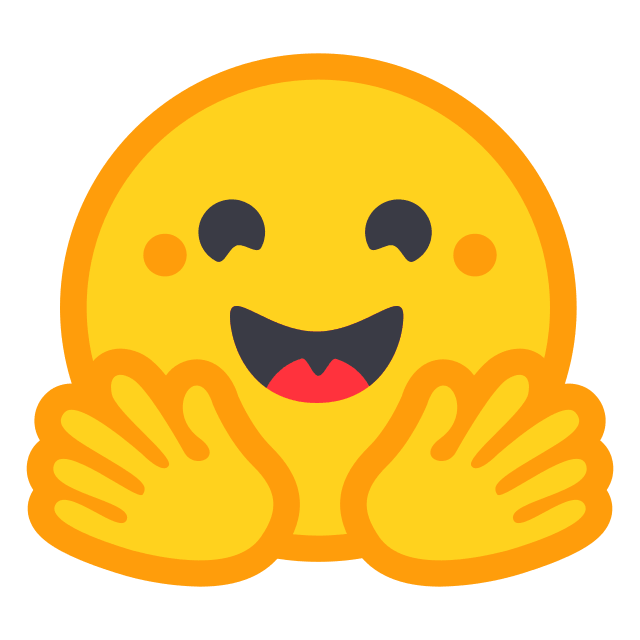}}}
\newcommand{\ghlogo}{\raisebox{-0.2em}{\includegraphics[height=1em]{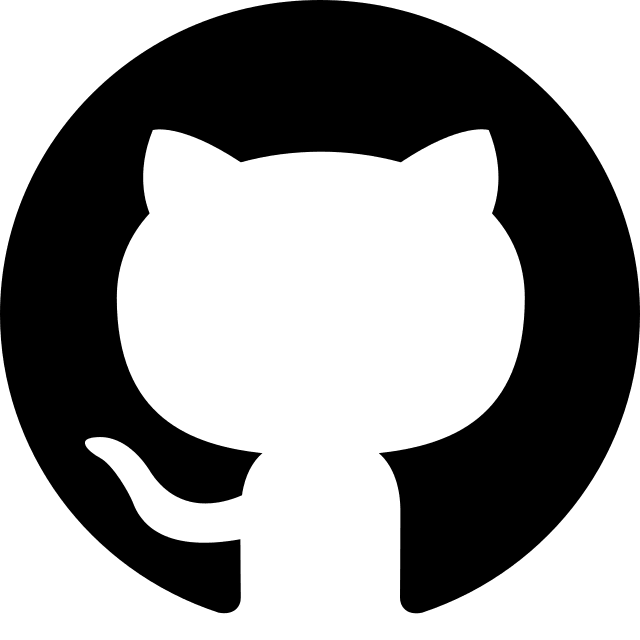}}}

\usepackage[utf8]{inputenc} 
\usepackage[T1]{fontenc}    
\usepackage{microtype}      
\usepackage{xcolor}         

\usepackage{amsmath,amsfonts,amssymb,mathtools} 
\usepackage{nicefrac}       

\usepackage[bookmarks,colorlinks=true,pagebackref=true,backref=page]{hyperref} 
\usepackage[capitalize]{cleveref} 
\usepackage{url}            

\usepackage{balance}        
\usepackage{hyphenat}       
\usepackage{stfloats}       
\usepackage{verbatim}       

\usepackage{booktabs}       
\usepackage{multirow}       
\usepackage{array}          
\usepackage{rotating}

\usepackage{graphicx}       
\usepackage[style=base, tableposition=top]{caption} 
\usepackage[caption=false,font=normalsize,labelfont=sf,textfont=sf]{subfig} 
\usepackage{tikz}           
\usetikzlibrary{patterns, positioning, arrows, bayesnet, calc} 
\usepackage{tikz-cd}        
\usepackage{pgfplots}       
\usepgfplotslibrary{patchplots}
\pgfplotsset{compat=1.15}   
\usepackage[most]{tcolorbox}      
\usepackage{wrapfig}

\usepackage{algorithm}
\usepackage{algorithmic}

\def \nclusters{\kappa}
\def \nsamples{\varrho}
\def \cluster{\mathcal{C}}

\title{\textsc{Semasia}: A Large-Scale Dataset of Semantically Structured Latent Representations} 

\author{%
  Mario Edoardo Pandolfo\thanks{Equal Contribution.}\,\,\,\thanks{\emph{Dept. Computer, Control and Management Engineering}, Sapienza University of Rome, Rome, Italy.}\,\,\,\thanks{\emph{National Inter-University Consortium for Telecommunications (CNIT)}, Parma, Italy.}\\
  \texttt{marioedoardo.pandolfo@uniroma1.it} \\
  \And
  Enrico Grimaldi\footnotemark[1]\,\,\,\footnotemark[2]\,\,\,\footnotemark[3]\\
  \texttt{enrico.grimaldi@uniroma1.it}
  \And
  Lorenzo Marinucci\thanks{\emph{Dept. of Statistical Sciences}, Sapienza University of Rome, Rome, Italy.}\\
  \texttt{l.marinucci@uniroma1.it}
  \And
  Leonardo Di Nino\thanks{\emph{Dept. of Information Engineering, Electronics, and Telecommunications}, Sapienza University of Rome, Rome, Italy.}\,\,\,\footnotemark[3]\\
  \texttt{leonardo.dinino@uniroma1.it}
  \And
  Simone Fiorellino\footnotemark[2]\\
  \texttt{simone.fiorellino@uniroma1.it}
  \And
  Sergio Barbarossa\footnotemark[5]\,\,\,\footnotemark[3]\\
  \texttt{sergio.barbarossa@uniroma1.it}
  \And
  Paolo Di Lorenzo\footnotemark[5]\,\,\,\footnotemark[3]\\
  \texttt{paolo.dilorenzo@uniroma1.it}
}

\begin{document}

\maketitle

\doparttoc
\faketableofcontents

\begin{abstract}
Latent representations learned by neural networks often exhibit semantic structure, where concept similarity is reflected by geometric proximity in embedding space. However, comparing such spaces across models remains difficult: changes in architecture, pretraining data, objective, or random seed can yield embeddings with similar content but incompatible geometry. This latent space alignment problem is central to interpretability, transfer and multimodal learning, federated systems, and semantic communication; however, progress remains limited by the lack of large-scale, model-diverse, and metadata-rich benchmarks.\\
To address this gap, we introduce \textsc{Semasia}, a large-scale collection of latent representations extracted from approximately 1,700 pretrained vision models across eight standard image-classification benchmarks. \textsc{Semasia} pairs embeddings with structured metadata describing architectures, training regimes, pretraining sources, and model scale. We demonstrate three applications of the resource. First, we analyze the conceptual organization of individual latent spaces, showing consistent prototype-like clustering and hierarchical semantic neighborhoods across models and datasets.
Second, we benchmark supervised alignment mappings between latent spaces using reconstruction error and downstream task performance. Third, we perform a large-scale regression analysis of how pretraining-data complexity, specialization, transfer learning, augmentation, and model scale relate to geometric and probing properties of embeddings. By coupling representational scale with standardized metadata, \textsc{Semasia} provides a reproducible foundation for studying latent geometry, evaluating alignment methods, and developing next-generation heterogeneous and interoperable AI systems.

\small Dataset: \hflogo \, \url{https://huggingface.co/collections/spaicom-lab/semasia}.

\small Code: \ghlogo\,\url{https://github.com/SPAICOM/semasia-datasets}.
\end{abstract}

\section{Introduction}

\begin{figure}[t]
    \centering
    \includegraphics[width=\textwidth,trim={0 12cm 0 0}, clip]{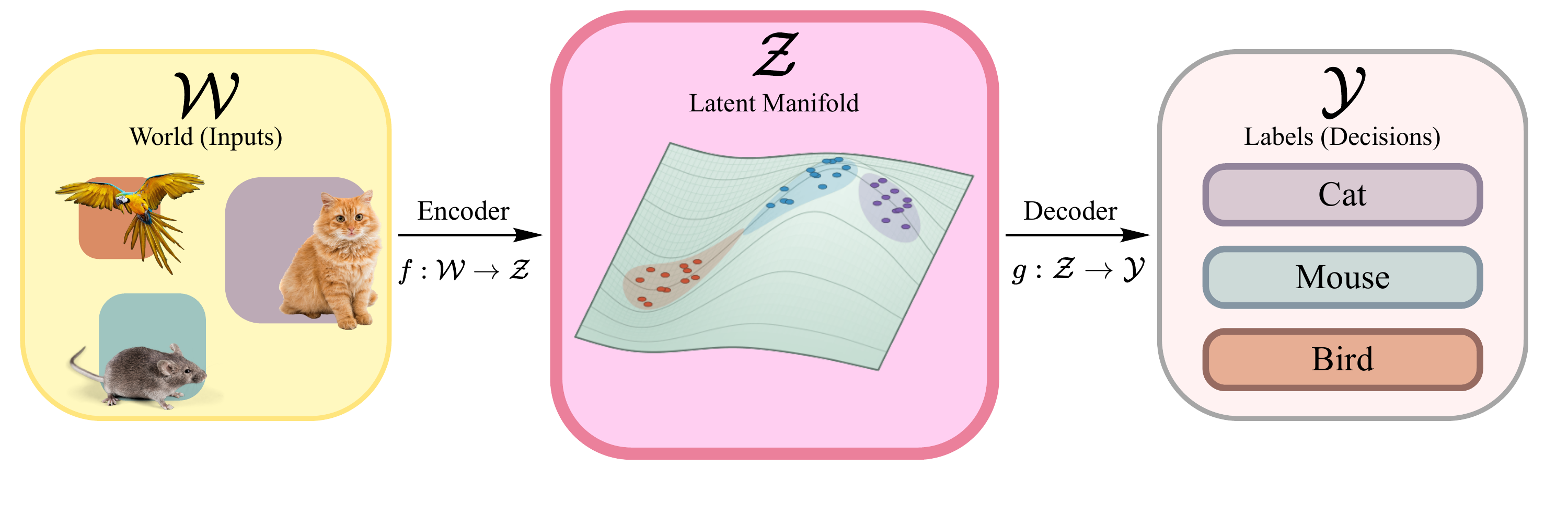}
    \caption{Illustration of the semiotic pipeline underlying modern neural models: raw perceptual inputs from the world $\mathcal{W}$ (e.g., images of animals) are mapped by an encoder $f: \mathcal{W} \rightarrow \mathcal{Z}$ into a latent manifold $\mathcal{Z}$, where inputs are organized into geometrically structured semantic clusters. A decoder $g: \mathcal{Z} \rightarrow \mathcal{Y}$ then projects these abstract representations onto a discrete label space $\mathcal{Y}$, yielding the final decision. This architecture mirrors the semiotic distinction between \textit{signifier} (the raw input) and \textit{signified} (the latent encoding), and instantiates G\"{a}rdenfors' conceptual spaces framework, in which concepts correspond to regions of a continuous, geometrically meaningful space.}
\label{fig:semiotic_pipeline}
\end{figure}
\vspace{-4pt}

The nature of reality and the mechanisms through which we perceive and represent it have long been central themes in philosophy, cognitive science, and, more recently, machine learning. A decisive turning point in this discourse came with the advent of neural models such as the perceptron~\citep{rosenblatt1958perceptron}, which introduced the idea that machines could ingest raw sensory inputs and automatically extract \textit{meaningful} features. This paradigm, later expanded into modern deep learning~\citep{lecun2015deep,rumelhart1986learning}, has enabled the construction of highly expressive models that can be directly identified with a semiotic process~\cite{eco1979theory}, capable of mapping observations from the physical world---the \textit{signifier}---into abstract representations---the \textit{signified}---that support complex reasoning and decision-making.
From the perspective of statistical learning, this corresponds to learning a mapping from an input domain rooted in perception, $\mathcal{W}$, to an output space $\mathcal{Y}$, through intermediate latent representations in $\mathcal{Z}$, as illustrated in Figure~\ref{fig:semiotic_pipeline}. These latent variables can be interpreted as semantic encodings: abstractions that progressively detach from the syntactic structure of the input and retain only the information relevant for downstream tasks. This view connects naturally to geometric theories of meaning. In cognitive science, G\"ardenfors' \textit{conceptual spaces} framework~\citep{gardenfors2000conceptual,gardenfors2014geometry} models concepts as regions in a geometrically structured space, where dimensions correspond to interpretable semantic features. Similarly, Osgood's \textit{semantic differential}~\citep{osgood1957measurement} represents meaning as a point in a low-dimensional evaluative space. Modern machine learning instantiates these ideas through embeddings~\citep{mikolov2013efficient,pennington2014glove}, where semantic similarity is reflected by geometric proximity, and through architectures that explicitly manipulate representations in continuous latent spaces.

Across architectures, latent representations emerge in different forms. In feedforward deep neural networks, they are organized hierarchically across layers, with increasing levels of abstraction~\citep{bengio2013representation}. In autoencoders~\citep{kingma2014auto} and U-Net-like architectures~\citep{ronneberger2015unet}, a \textit{semantic bottleneck} explicitly compresses the input into a compact representation before reconstruction or prediction. In contrast, non-hierarchical models such as Hopfield networks~\citep{hopfield1982neural} encode information in the global activation state of a recurrent system, where the latent representation coincides with the network's energy minima. Despite these differences, a common principle emerges: latent spaces provide a structured representation of meaning that abstracts away from raw sensory inputs.

Understanding the geometry of these latent spaces has become a central challenge in representation learning. Two complementary research directions have emerged in this context. On one hand, \textit{representation analysis} methods~\citep{raghu2017svcca,morcos2018insights,kornblith2019similarity} study the global structure of latent spaces and quantify similarities across models. On the other hand, \textit{mechanistic interpretability}~\citep{bereskamechanistic} seeks to reverse-engineer the computations that give rise to these representations. Together, these approaches aim to uncover how neural networks encode and manipulate information.

A striking observation from this line of work is that representations learned by different models often exhibit convergent geometric structure. This has led to the formulation of the \textit{Platonic Representation Hypothesis}~\citep{huh2024platonic}, which posits that sufficiently powerful models approximate a shared statistical representation of reality. Empirically, this manifests as alignment in the similarity structure (kernel) of latent spaces across models and modalities. However, such convergence is inherently imperfect: unlike humans, whose representations are shaped by a strong pragmatic need to communicate with other social agents \citep{zeman1977peirce, watzlawick2011pragmatics}, the task-tailored latent space of a neural model is subject to no such constraint. As a consequence, the semantic codes produced by a neural network are not 
directly comparable: even minor sources of stochasticity---such as weight initialization, optimization dynamics, or data shuffling---introduce variability, while differences in architecture, modality, and training data further amplify discrepancies between latent spaces, yielding representations that are equivalent yet misaligned~\citep{javidnia2026gauge}.

These discrepancies become critical in settings where representations themselves are exchanged, such as \textit{semantic communications}~\citep{shi2021new,gunduz2022beyond,strinati2024goal}, where latent codes act as communication units~\citep{xie2021deep}. In this regime, misaligned latent spaces induce \textit{semantic noise}~\citep{sana2023semantic,luo2022semantic}, hindering mutual understanding and limiting interoperability across agents. This challenge is especially acute in AI-native 6G systems, where heterogeneous models must communicate directly through learned representations. More generally, any paradigm operating in representation space—including transfer learning, multimodal and multitask modeling~\citep{radford2021clip,girdhar2023imagebind,cicchetti2025gramian}, federated learning~\citep{tan2022fedproto,yang2023fedfed,benissaid2026fedmuscle,badi2026comfed}, and multi-agent systems—faces the same fundamental issue: without proper \textit{latent space alignment}, representations remain semantically incompatible.

\textbf{Contributions and Impact.}
Our contributions are threefold. First, we release a large-scale, standardized dataset of latent representations that captures the expressivity of SoTA vision models, highlighting their capability of extracting concepts at different granularities. Second, we navigate the diversity of modern neural architectures and datasets by comparing latent representations and studying the effect of specific training and modeling choices on latent spaces. Third, we establish a unified benchmark for evaluating latent space alignment methods under realistic and heterogeneous conditions.

We believe that \textsc{Semasia} provides a crucial step toward a principled understanding of semantic representations in neural networks. It opens new avenues for studying the geometry of meaning, advancing alignment methodologies, and supporting the development of next-generation AI communication systems capable of robust and meaningful interaction.
Furthermore, this dataset can be used not only for benchmarking in semantic communication setups, but also as a basis for fundamental studies into the implementation of innovative techniques such as parameter-efficient fine-tuning, transfer learning, and model steering, where the geometry of the latent space and the analysis of its structure are becoming crucial~\citep{hu2022lora,zou2023representation,grant2026gluing}.
Finally, we believe that the analyses presented in the paper advance the state of the art in the field of explainable AI~\citep{zhu2024towards}, and may inspire the design of new multimodal and multitask architectures, as well as potential studies of latent dynamics~\citep{fumero2025navigating} at the core of world models~\citep{lecun2022path}.


\section{The \textsc{Semasia} Dataset}
\label{sec:dataset}

\textsc{Semasia} is a large-scale collection of latent representations extracted from state-of-the-art neural vision models available in the \texttt{timm} library~\citep{rw2019timm}. Each representation is obtained by feeding images from a standard computer vision benchmark into a pretrained model in inference mode, and recording the vector of neural activations at a designated layer. Concretely, we extract activations from the last layer immediately preceding the classification or task-specific MLP decoder, effectively repurposing \texttt{timm} models as semantic feature extractors rather than classifiers. The rationale behind this choice is discussed in Section~\ref{sec:lse} and Appendix \ref{app:cutting-point}.

\subsection{Models and Benchmarks}
\label{sec:models-benchmarks}

The \texttt{timm} library hosts approximately $1{,}700$ pretrained vision models. In this work, we restrict our analysis to a subset of $1697$ architectures, excluding models on the order of $10^{10}$ parameters due to computational constraints imposed by the available hardware. The selected models span a broad spectrum of design choices, including convolutional networks, vision transformers, hybrid architectures, and self-supervised backbones, providing a representative cross-section of modern vision encoders. We release \textsc{Semasia} on Hugging Face as a dataset collection, with one semantic dataset per benchmark. The current release covers eight widely adopted vision benchmarks: CIFAR-10 \citep{Krizhevsky2009}, CIFAR-100 \citep{Krizhevsky2009}, MNIST \citep{LeCun1998}, Fashion-MNIST \citep{Xiao2017}, Oxford Flowers \citep{Nilsback2008}, ImageNet-1k \citep{Deng2009, Russakovsky2015}, Tiny ImageNet \citep{Le2015} and CelebA \citep{Liu2015}. Source datasets are obtained from their respective Hugging Face repositories, and latent representations are extracted on one or more of the default splits provided by the dataset authors, as summarized in Table~\ref{tab:datasets}. Additional details on the benchmarking datasets used are reported in Appendix~\ref{app:datasets}.

\subsection{Data Format and Organization}
\label{sec:data-format}

\begin{table}[t]
  \centering
  \caption{Summary of the benchmark datasets included in the \textsc{Semasia} collection.
    For each dataset and split we report the number of \texttt{timm} models from which latent representations were extracted, the number of source examples per model, and the total number of rows in the corresponding Parquet file (i.e., the number of (input, model) pairs).}
  \label{tab:datasets}
  \vspace{0.5em}
  \small
  \setlength{\tabcolsep}{5pt}
  \begin{tabular}{llrrr}
    \textbf{\textcolor{darkviolet}{Dataset}} & \textbf{\#\,Models} & \textbf{Split} & \textbf{Raw examples} & \textbf{Total Rows} \\
    \textbf{\href{https://huggingface.co/collections/spaicom-lab/semasia}{\textcolor{darkviolet}{\hflogo\,\texttt{semasia-collection}}}} & & & & \\
    \hline
    \multirow{3}{*}{\href{https://huggingface.co/datasets/spaicom-lab/semasia-celeba}{\textcolor{datasetpurple}{\texttt{semasia-celeba}}}}
      & \multirow{3}{*}{$1{,}697$} & train      & $100{,}000$ & $169{,}700{,}000$ \\
      &                            & valid      & $19{,}867$  & $33{,}714{,}299$  \\
      &                            & test       & $19{,}962$  & $33{,}875{,}514$  \\
    \midrule
    \multirow{2}{*}{\href{https://huggingface.co/datasets/spaicom-lab/semasia-cifar10}
    {\textcolor{datasetpurple}{\texttt{semasia-cifar10}}}}
      & \multirow{2}{*}{$1{,}697$} & train      & $50{,}000$  & $84{,}850{,}000$  \\
      &                            & test       & $10{,}000$  & $16{,}970{,}000$  \\
    \midrule
    \multirow{2}{*}{\href{https://huggingface.co/datasets/spaicom-lab/semasia-cifar100}
    {\textcolor{datasetpurple}{\texttt{semasia-cifar100}}}}
      & \multirow{2}{*}{$1{,}697$} & train      & $50{,}000$  & $84{,}850{,}000$  \\
      &                            & test       & $10{,}000$  & $16{,}970{,}000$  \\
    \midrule
    \multirow{2}{*}{\href{https://huggingface.co/datasets/spaicom-lab/semasia-fashion_mnist}{\textcolor{datasetpurple}{\texttt{semasia-fashion\_mnist}}}}
      & \multirow{2}{*}{$1{,}697$} & train      & $60{,}000$  & $101{,}820{,}000$ \\
      &                            & test       & $10{,}000$  & $16{,}970{,}000$  \\
    \midrule
    \multirow{2}{*}{\href{https://huggingface.co/datasets/spaicom-lab/semasia-mnist}{\textcolor{datasetpurple}{\texttt{semasia-mnist}}}}
      & \multirow{2}{*}{$1{,}697$} & train      & $60{,}000$  & $101{,}820{,}000$ \\
      &                            & test       & $10{,}000$  & $16{,}970{,}000$  \\
    \midrule
    \multirow{2}{*}{\href{https://huggingface.co/datasets/spaicom-lab/semasia-oxford-flowers}{\textcolor{datasetpurple}{\texttt{semasia-oxford-flowers}}}}
      & \multirow{2}{*}{$1{,}697$} & train      & $7{,}169$   & $12{,}180{,}131$  \\
      &                            & test       & $1{,}020$   & $1{,}732{,}980$   \\
    \midrule
    \multirow{2}{*}{\href{https://huggingface.co/datasets/spaicom-lab/semasia-tiny-imagenet}{\textcolor{datasetpurple}{\texttt{semasia-tiny-imagenet}}}}
      & \multirow{2}{*}{$1{,}697$} & train      & $100{,}000$ & $169{,}700{,}000$ \\
      &                            & valid      & $10{,}000$  & $16{,}970{,}000$  \\
    \midrule
    \multirow{2}{*}{\href{https://huggingface.co/datasets/spaicom-lab/semasia-imagenet-1k}{\textcolor{datasetpurple}{\texttt{semasia-imagenet-1k}}}}
      & \multirow{2}{*}{$1{,}697$} & validation & $50{,}000$  & $84{,}850{,}000$  \\
      &                            & test       & $100{,}000$ & $169{,}700{,}000$ \\
    \bottomrule
  \end{tabular}
\end{table}

Data are stored in tabular form, with one Parquet file per neural model and per benchmark in the world space~$\mathcal{W}$ (e.g., \texttt{semasia-cifar10}, \texttt{semasia-mnist}). Each file has one row per input example, corresponding to a pair (input $i$, model $k$), and describes the latent representation of the $i$-th sample extracted from model $k$.

Each row contains: (i) \texttt{id} (\texttt{uint32}), a unique identifier linking representations of the same input across models, enabling exact correspondence for alignment studies; (ii) one or more label columns, inherited from the original benchmark and task-dependent, supporting analyses such as linear probing, representation quality evaluation, and alignment or semantic communication benchmarking (e.g., mapping representations from a model~$A$ to a decoder of a model~$B$), with dataset-specific structure (\texttt{semasia-cifar10}: single integer label; \texttt{semasia-celeba}: multiple binary attributes; see Appendix~\ref{app:datasets}); (iii) \texttt{model\_name} (\texttt{string}), the identifier in the \texttt{timm} library; and (iv) \texttt{embedding} (\texttt{array} of shape equal to the latent dimension of the corresponding model), the latent representation extracted as described in Section~\ref{sec:lse}.

\subsection{Model Registry}
\label{sec:model-registry}

The dataset is accompanied by a \textit{model registry} that records metadata for each architecture used to extract latent representations, including identity, architectural characteristics (e.g., family, depth, width, input resolution), pretraining provenance, and capacity descriptors such as parameter count and latent dimensionality. This information enables controlled analyses of how design choices, scale, and training affect latent space geometry. A full description is provided in Appendix~\ref{app:registry}. The architectural heterogeneity documented by the model registry makes \textsc{Semasia} a meaningful testbed for alignment, as latent spaces are not directly comparable across models.

\begin{figure}[t]
    \centering
    \includegraphics[width=\textwidth,trim={0 0 3.55cm 0}, clip]{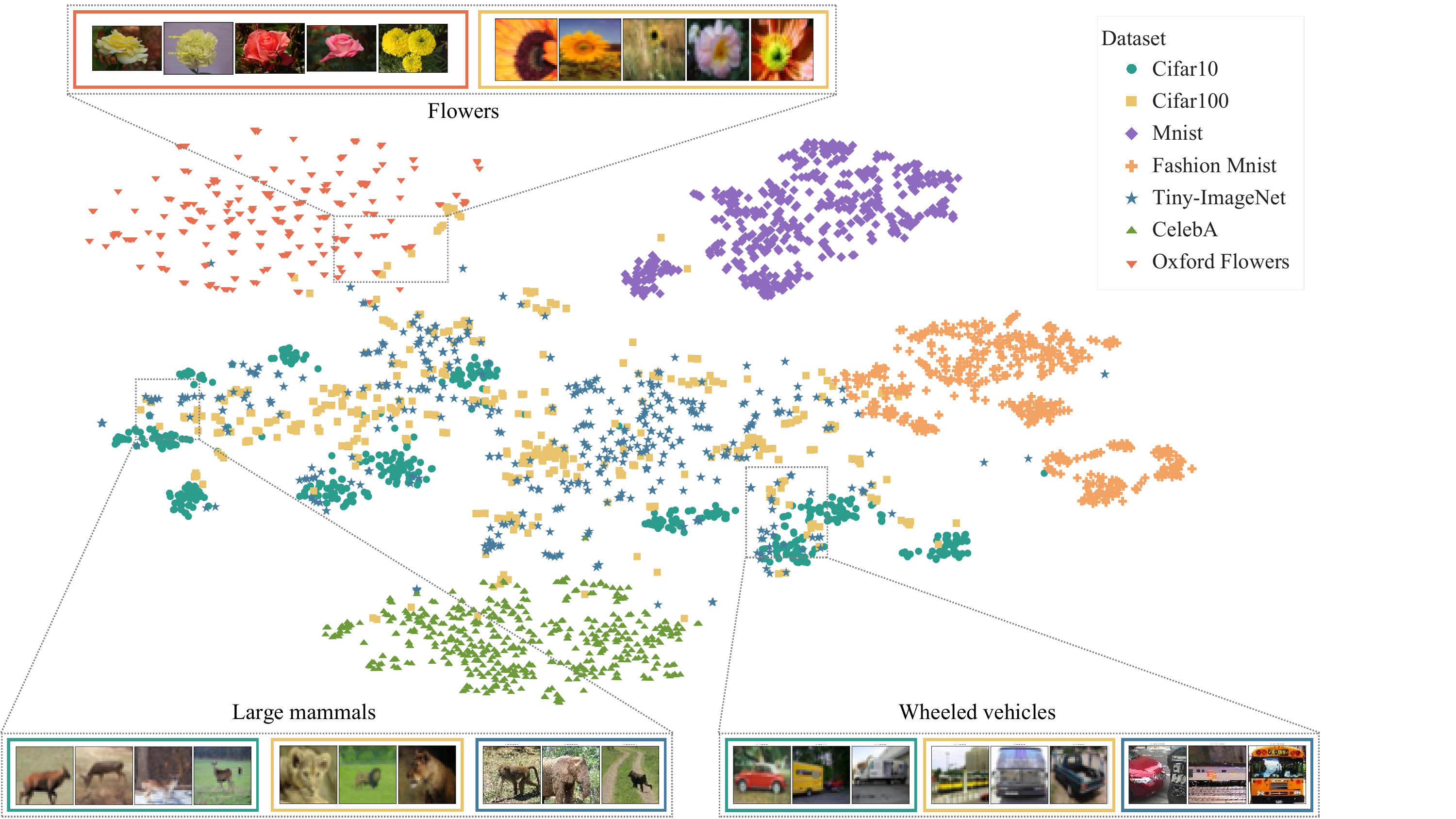}
\caption{Two-dimensional t-SNE projection of the \texttt{aimv2\_1b\_patch14\_224.apple\_pt} 2048-dimensional latent space, populated with samples from seven \textsc{Semasia} benchmarks. Each benchmark forms its own cluster, but semantically overlapping concepts collapse onto shared neighborhoods regardless of source: flower images from Oxford~Flowers and CIFAR-100 occupy the same region, as do large mammals and vehicles drawn from CIFAR-10, CIFAR-100, and Tiny-ImageNet (highlighted with the related origin dataset color).}
    \label{fig:tsne-main}
\end{figure}

\begin{figure}[t]
    \centering
    \includegraphics[width=\textwidth,trim={2.3cm 0 0 0}, clip]{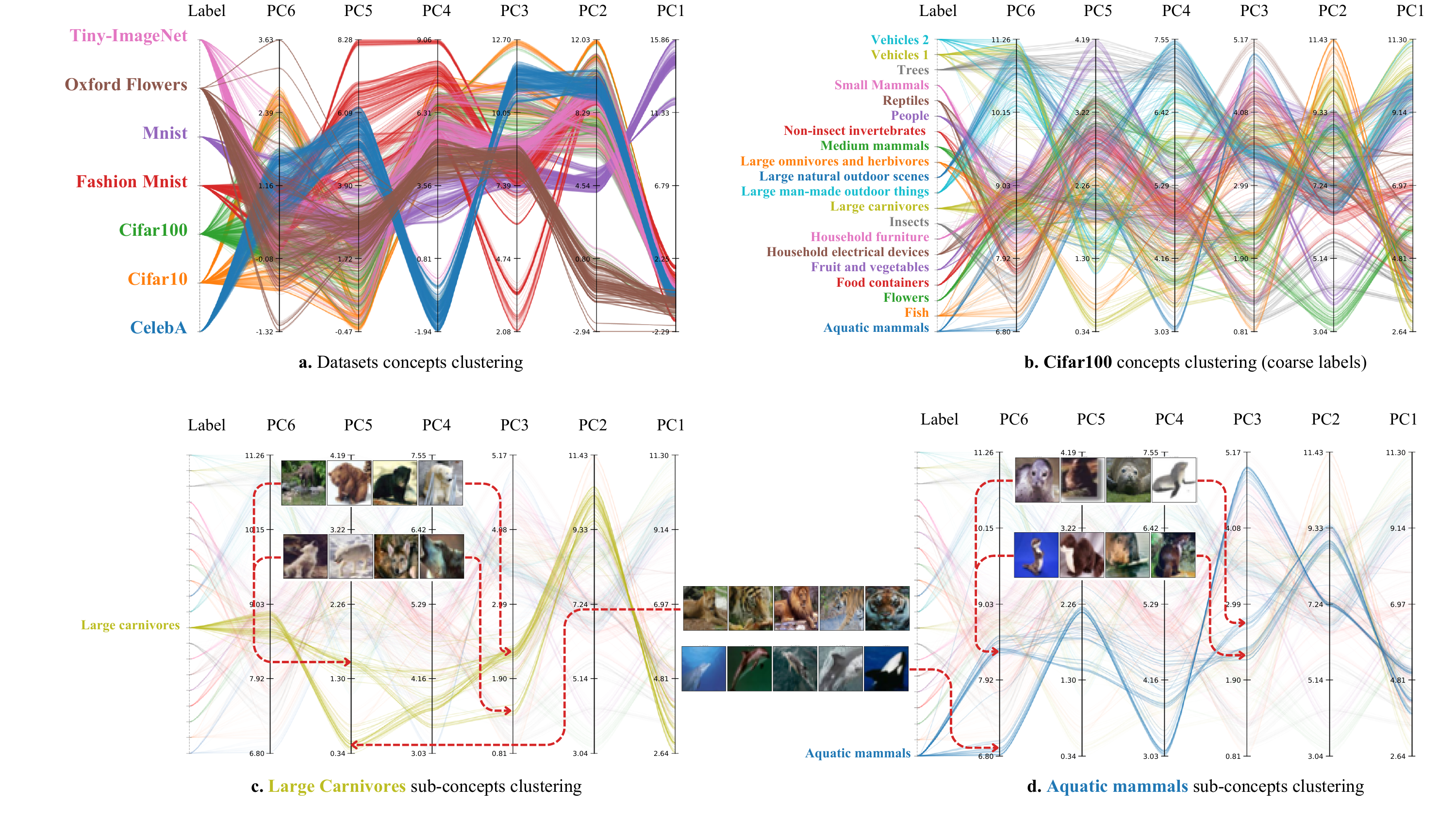}
    \caption{Concept clustering in the \texttt{aimv2\_1b\_patch14\_224.apple\_pt} latent space, projected along six principal components of UMAP. Each axis encodes a latent feature, and examples organize along it according to how the feature is realized. \textbf{(a)}~Distribution of latent representations from the full \textsc{Semasia} collection, showing how each benchmark forms its own cluster (complementing Figure~\ref{fig:tsne-main}). \textbf{(b)}~Same projection restricted to CIFAR-100, revealing how its classes distribute along the principal axes. \textbf{(c, d)}~Zoom on large carnivorous and aquatic mammals, respectively. In (c), PC~5 separates felines from canids/bears, and PC~4 further splits canids from bears. In (d), the first axis separates cetaceans from furred mammals, while PC~3 distinguishes rodents/mustelids from pinnipeds within the aquatic furred group.}
    \label{fig:concepts}
    \vspace{-6pt}
\end{figure}

\subsection{Latent Space Extraction}
\label{sec:lse}

\paragraph{Extraction protocol.}
For each model in the registry and each benchmark dataset, we feed every raw input through the network in inference mode and record the vector of neural activations at a designated \textit{cutting point}. Each input is preprocessed according to the default pipeline specified by the \texttt{timm} maintainers for the target model, typically including resizing, padding, and normalization to match the expected input dimensionality and statistics. The cutting point is chosen as the last layer immediately preceding the task-specific decoder, i.e., the head used at training time for the supervised objective. This choice is principled: while not every architecture exposes an explicit semantic bottleneck, most modern vision models exhibit a hierarchical structure in which early layers extract low-level, syntactic features tied to the geometry of the input, while deeper layers progressively shed input-specific structure in favor of task-relevant semantic content~\citep{bengio2013representation}. Cutting one layer before the head, therefore, yields a representation that has largely abandoned the geometry of the input space but has not yet collapsed onto the discrete output space~$\mathcal{Y}$ of the supervised task. Empirical evidence supporting this choice is provided in Appendix~\ref{app:cutting-point}.

\textbf{Latent spaces as point clouds.}
Recording one latent representation per input induces, for each (model, dataset) pair, a point cloud in $\mathcal{Z}$ that we treat as an empirical sample of the model's latent space. By aggregating across the training splits of all benchmark datasets, each \textsc{Semasia} entry covers the latent space at scale and across heterogeneous semantic domains, providing a comprehensive substrate for downstream geometric and topological analyses. Figure~\ref{fig:tsne-main} visualizes such a point cloud for a single model, \texttt{aimv2\_1b\_patch14\_224.apple\_pt} (model details in \citep{fini2025multimodal}), aggregated over the training splits of seven benchmarks and projected to two dimensions via t-SNE ~\citep{van2008visualizing}.

\textbf{Geometry of meaning.}
The projection in Figure~\ref{fig:tsne-main} provides a concrete realization of the cognitive-science intuition that semantic content can be organized as a geometric space in which proximity encodes similarity~\citep{gardenfors2000conceptual,osgood1957measurement, wakhloo2026neural}. Three observations are particularly noteworthy. First, semantically equivalent concepts collapse onto overlapping regions regardless of their dataset of origin: instances of the concept \textit{flower} drawn from CIFAR-10 and from Oxford Flowers occupy the same neighborhood in the projection, despite the model never having been directly trained on either benchmark. Second, individual concepts manifest as compact, well-localized clusters whose centroids can be read as concept prototypes, in line with recent prototype-based interpretations of neural representations~\citep{fiorellino2025frame}. Third, the latent space is organized hierarchically: parallel-coordinates plots~\citep{inselberg1985plane} along the principal UMAP ~\citep{mcinnes2018umap} axes in Figure~\ref{fig:concepts}, reveal that concepts cluster at multiple levels of granularity, with directions in the projection encoding interpretable semantic axes. For instance, along one axis felines, canids, and bears are progressively separated, while along another, canids and bears are further disentangled from one another. The model thus appears to extract a semantics that supports discrimination between classes and subclasses of benchmarks on which it was never explicitly trained. A deeper exploration of concept clustering at varying granularities for the example model across different \textsc{Semasia} benchmarks is reported in Appendix~\ref{app:concept-clustering}.
The above analysis describes the latent space of a \textit{single} model. A central question of the semantic-alignment literature, however, concerns whether latent spaces produced by \textit{different} models are directly comparable, and whether they are semantically equivalent up to a structured transformation. Investigating this question, both within and across architectural families, is the focus of Sections~\ref{sec:align} and~\ref{sec:stat_analysis}.




\subsection{Extensibility and Compatibility}
\label{sec:extensibility}

The collection can be extended along two axes: applying the latent space extraction pipeline to new datasets or modalities (e.g., audio, text, video), or analyzing latent spaces of newly proposed architectures on fixed benchmarks. This design enables systematic studies of alignment across heterogeneous models, data modalities, and task-tailored representations. A key motivation is to investigate the geometric and topological structure of latent spaces, with topology-based methods offering insights into semantic alignment and disentanglement. \textsc{Semasia} is already integrated in the open-source framework \textsc{Topobench}~\citep{bernardez2026topological}, allowing direct evaluation of classical and deep learning methods for point cloud analysis, manifold learning, and topological inference.
\section{\textsc{Semasia} in Action: From Alignment to Statistical Analysis}
\label{sec:use_cases}

This section illustrates how \textsc{Semasia} provides a controlled benchmark for latent space alignment methods and enables the systematic study of semantic mismatch. In section \ref{sec:align}, we focus on CIFAR-10 representations from a small set of selected model pairs, and test methods from the literature on latent space alignment; the protocol extends naturally to the full collection.
In section \ref{sec:stat_analysis}, we instead leverage the full \textsc{Semasia} arsenal to conduct what is, to our knowledge, the first large-scale regression-based analysis of latent space geometry across vision models.

\subsection{Semantic Alignment}
\label{sec:align}

\begin{figure}[t]
    \centering
    \includegraphics[width=\textwidth]{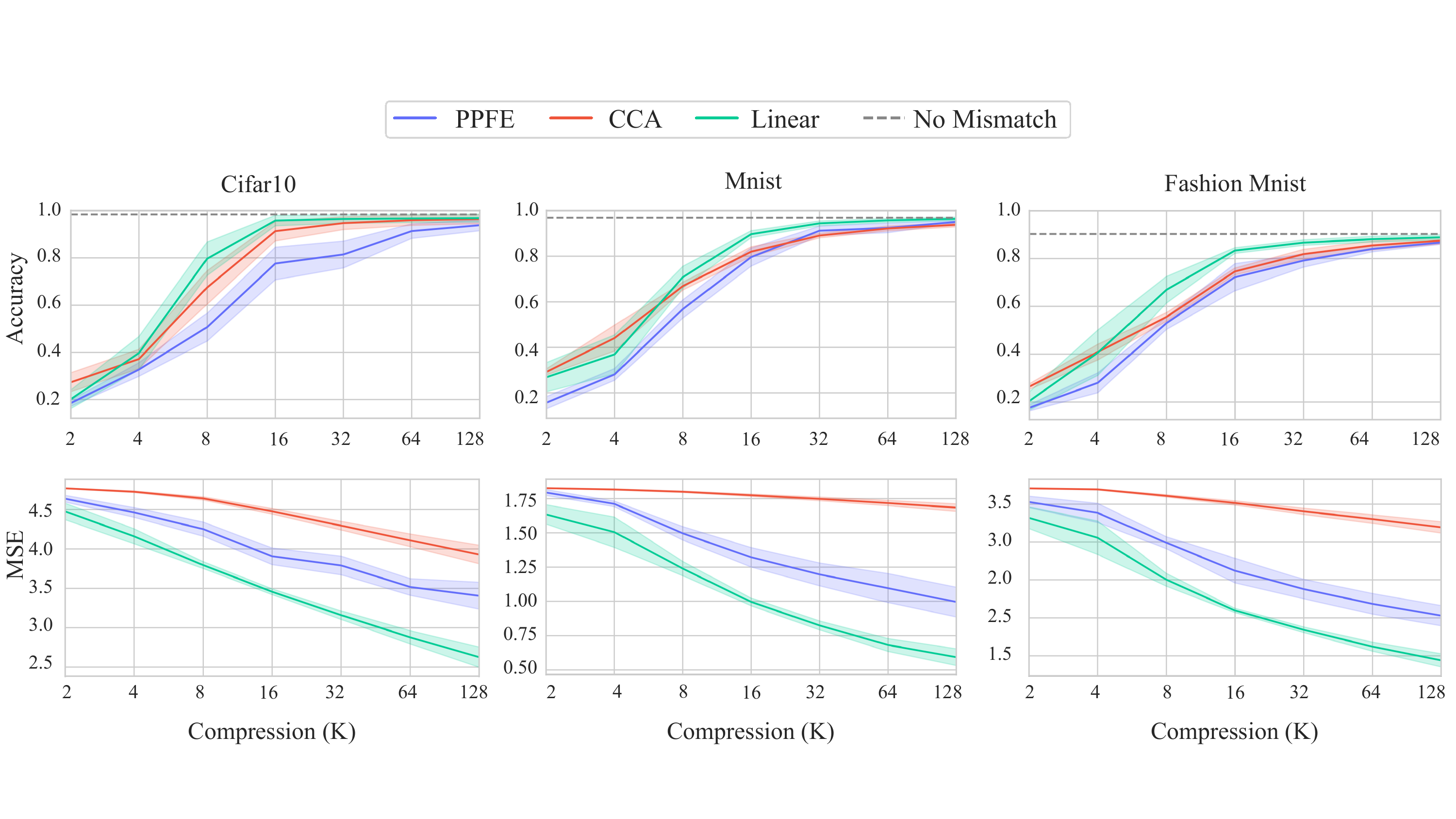}
\caption{Comparison of three supervised alignment methods on every model pair from Figure~\ref{fig:pca}: \textsc{Linear} (the Eigen-$K$ baseline from~\citealp{pandolfo2025latent}), \textsc{PPFE}~\citep{fiorellino2025frame}, and Canonical Correlation Analysis (\textsc{CCA})~\citep{raghu2017svcca}. The $x$-axis reports the number of non-zero components $K$ retained by the alignment map (i.e., the active latent dimensions); the $y$-axis reports downstream task accuracy after communication (top) and mean squared error of latent reconstruction (bottom). Method specifications and metric definitions are detailed in Appendix~\ref{app:alignment}.}
    \label{fig:alignment}
\end{figure}

Latent spaces produced by different models exhibit structured forms of semantic mismatch. These discrepancies can be characterized geometrically, either through comparisons of latent bases (Appendix~\ref{app:latent_bases}) or via correspondence of representative concepts (Appendix~\ref{app:compare_concepts}). For completeness, we present a detailed empirical analysis of these phenomena in the corresponding appendices, focusing on a set of carefully selected representative models chosen to isolate specific sources of heterogeneity that we examine more broadly in Section~\ref{sec:stat_analysis}. Together, these results, spanning both basis-level comparisons and concept-level correspondences, highlight the need for explicit alignment mappings.

Figure~\ref{fig:alignment} compares three supervised alignment methods—a linear map (Linear)~\citep{pandolfo2025latent}, a prototype-anchor projection (PPFE)~\citep{fiorellino2025frame}, and Canonical Correlation Analysis (CCA)~\citep{raghu2017svcca}—on ViT and AiMV2 model pairs. Each curve traces performance as a function of the number of non-zero components $K$, i.e., the number of latent dimensions retained by the alignment map. Lower values of $K$ correspond to more aggressive compression and thus to a stricter test of how much task-relevant semantic content survives the alignment. We report two complementary metrics: latent reconstruction quality and downstream task accuracy in a semantic-communication setting. Full experimental details are provided in Appendix~\ref{app:alignment}. Across all model pairs and across the full range of compression factors, Linear consistently dominates both PPFE and CCA on both metrics. The gap is most pronounced at low values of $K$, where prototype-based and CCA-based alignments degrade sharply while Linear retains most of the downstream accuracy and reconstruction fidelity. These results highlight the role of \textsc{Semasia} as a benchmarking resource: rather than committing to an alignment method on theoretical grounds, practitioners can run lightweight comparisons on relevant latent space pairs and select the best-performing strategy for the heterogeneity regime at hand. The same protocol extends naturally to new alignment methods as they are proposed.

\subsection{Statistical Analysis of Embedding Geometry}
\label{sec:stat_analysis}

\begin{figure}[t]
    \centering
    \includegraphics[width=\textwidth,trim={0 1.5cm 0 0}, clip]{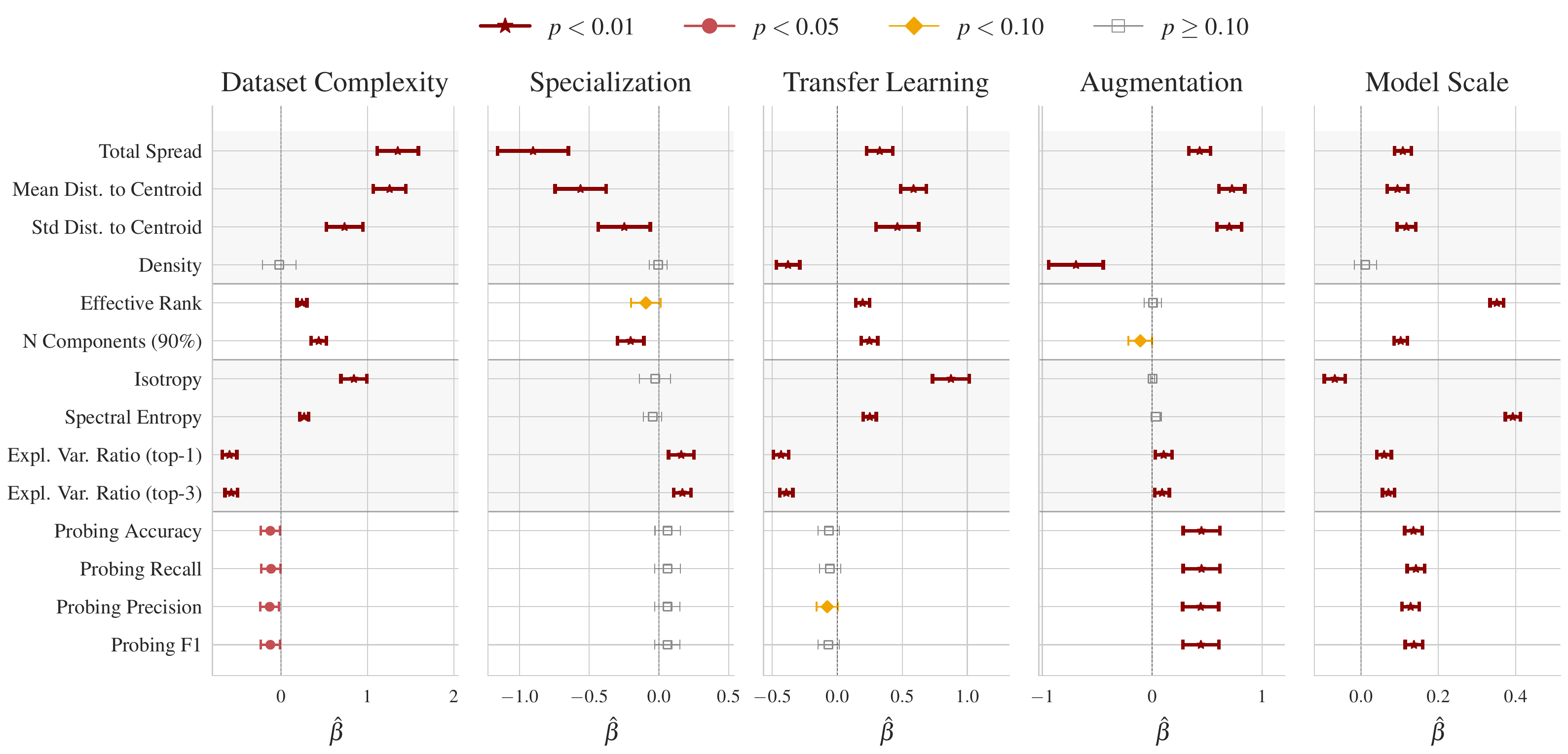}
    \caption{Forest plot of pooled OLS regression coefficients $\hat{\beta}$ for the embedding geometry and linear probing metrics across five pretraining conditions (see Table~\ref{tab:conditions} for condition definitions). Probing scores (accuracy, recall, precision, F1) are obtained via a least-squares linear probe. All $\hat{\beta}$ are expressed in units of the control-group standard deviation of each metric ($\sigma_{\text{control}}$, pooled across datasets and analysis types), yielding an effect-size measure analogous to Cohen's $d$. Observations are pooled across CIFAR-10, MNIST, Fashion-MNIST, and Oxford Flowers.}
    \label{fig:forest}
    \vspace{-4pt}
\end{figure}

To demonstrate the analytical potential of SEMASIA, we regress fourteen geometric and probing metrics
(described in \ref{app:target_variables}), covering spread, intrinsic dimensionality, spectral
structure, and linear probing performance, against five pretraining conditions derived from the \textit{model registry} (Section \ref{sec:model-registry} and Appendix \ref{app:registry}). Each condition is
designed as a \textit{ceteris paribus} contrast, isolating a single factor while holding all others
fixed. We fit a pooled OLS regression with HC3-robust standard errors, controlling for architecture
family and evaluation dataset. Coefficients in Figure~\ref{fig:forest} are expressed in units of the
control-group standard deviation $\sigma_{\text{control}}$ (pooled across datasets and analysis
types), yielding an effect-size measure analogous to Cohen's $d$. The analysis spans CIFAR-10,
MNIST, Fashion-MNIST, and Oxford Flowers, yielding between 224 and 7,260 pooled observations per condition. Full condition definitions are given in
Table~\ref{tab:conditions} of Appendix~\ref{app:stat_regression}.

\textbf{Dataset Complexity} contrasts a smaller vs.\ a larger ImageNet variant, with architecture and augmentation fixed. Training on the richer dataset consistently expands the embedding space, increases effective rank, isotropy, and spectral entropy, and reduces explained variance of top components, indicating higher-dimensional, more uniformly distributed representations. The gain in effective rank reflects a broader activation of meaningful dimensions. Probing metrics are significantly negative, consistent with a semantic shift toward a distribution not directly aligned with the downstream classification task.
 
\textbf{Specialization} contrasts original large-scale pretraining with subsequent fine-tuning to a smaller dataset variant. Fine-tuning onto less rich data compresses the embedding space, reduces effective rank, isotropy, and spectral entropy, and concentrates variance along dominant directions, the geometric signature of catastrophic forgetting. Probing metrics are non-significant.
 
\textbf{Transfer Learning} contrasts native training on the target dataset with large-scale pretraining followed by fine-tuning to the same target, and tells the complementary story: pretrained models retain wider spaces, higher effective rank, and more uniform dimension utilisation. The preservation of effective rank indicates that large-scale pretraining leaves a lasting imprint on semantic coding capacity. Probing metrics are non-significant, again due to semantic shift.
 
\textbf{Augmentation} compares the same model and dataset with and without augmentation during pretraining. Augmentation expands distance-based spread and isotropy but does not increase effective rank; the number of components required to explain 90\% of variance even decreases slightly, indicating that augmentation redistributes variance more efficiently within already active dimensions rather than expanding semantic coding capacity. Unlike other conditions, probing metrics are significantly positive, confirming that a wider, more isotropic space aids linear classification when semantic structure is preserved.
 
\textbf{Model Scale} contrasts smaller vs.\ larger model variants within the same architectural family, dataset, and setup, and produces the strongest effects. Larger models yield wider spaces, higher effective rank, and significantly better probing performance, but with reduced isotropy: variance concentrates along dominant directions despite overall expansion. This combination distinguishes model scale from augmentation, which widens the space without increasing its semantic coding capacity. 

Dataset and architecture family fixed effects are jointly significant across all conditions ($p < 0.001$), confirming strong independent effects on embedding geometry.

Taken together, these results reveal a consistent \textbf{scale effect} across all conditions: training on a richer dataset expands the space and distributes semantic information across more orthogonal directions (\textbf{Dataset Complexity}), fine-tuning onto a less informative one compresses it (\textbf{Specialization}), and large-scale pretraining followed by fine-tuning preserves that capacity compared to native training (\textbf{Transfer Learning}). Increasing syntactic but not semantic variance through \textbf{Augmentation} widens the space without affecting intrinsic dimensionality, but simplifies the classification landscape for a fixed downstream probe. Finally, \textbf{Model Scale} both widens the space and substantially increases its semantic coding capacity, though with the notable signature of reduced isotropy, suggesting a richer but more anisotropic internal geometry. In Appendix~\ref{apps:graph-based-regression}, we conduct a statistical regression to assess whether different architecture families structure the latent space differently. Since model scale is not matched across architecture families, we rely on topological signatures, which are scale invariant by construction, to ensure that observed differences reflect architecture rather than model scale.
\section{Limitations, Discussion, and Conclusions}
\label{sec:conclusions}

\textsc{Semasia} provides the first opportunity to systematically study the geometry of latent spaces in vision models at scale, laying the foundation for a benchmark suite for semantic alignment that enables fast, reproducible comparison of latent spaces and both quantitative and qualitative analysis of semantic mismatch across heterogeneous configurations. Beyond semantic communication protocols, where alignment is a prerequisite for meaningful inter-agent exchange, the resource naturally extends to broader multi-agent and cooperative settings, including federated learning, model merging, and multimodal or multitask architectures. The statistical framework enabled by \textsc{Semasia} also provides the first regression-based evidence of how pretraining data complexity, specialization, transfer learning, augmentation, and model scale shape the geometry of vision embeddings. Such findings would have been difficult to obtain without a resource of this scope: pooling thousands of model--dataset observations within a unified regression framework, while controlling for architectural family and evaluation dataset, enables principled, hypothesis-driven studies of representation quality beyond task-specific performance metrics. Finally, all precomputed latent spaces are made publicly available, democratising access to large-scale representation benchmarking by eliminating the need for high-end GPU infrastructure, and decoupling latent space alignment research from hardware constraints by enabling researchers with modest computational resources to work with state-of-the-art architectures.

\textbf{Limitations and future directions.}
The current release focuses exclusively on image models, preventing direct cross-modal comparisons of latent geometries and the study of modality-specific deviations from a putative universal representation. Extending the collection to language, audio, and video models, and aligning their latent spaces on shared semantic content, is a natural next step for investigating the conditions under which the \textit{Platonic Representation Hypothesis} holds at scale. A second limitation is the use of a single extraction layer per model, fixed at the semantic bottleneck immediately preceding the task-specific head and evaluated only at the end of training. While this isolates the most semantically rich representation produced by each model, it leaves two complementary dimensions unexplored: \emph{depth} and \emph{time}. Interlayer analyses could clarify how representations evolve along the syntactic--semantic--pragmatic spectrum, while tracking latent spaces across training epochs could reveal how semantic geometry emerges during optimization, connecting \textsc{Semasia} to recent work on latent dynamics and world models. Another promising direction concerns the role of downstream objectives in shaping latent geometry. The modular design of \textsc{Semasia} is intended to support all these limitations as future extensions.

\textbf{Closing remarks.}
By coupling representational scale with structured metadata, \textsc{Semasia} treats latent representations not as opaque vectors but as measurable, comparable, and alignable geometric objects. We hope it will provide a shared substrate for advancing the study of meaning in artificial systems, developing new alignment methodologies, and ultimately enabling robust semantic communication among heterogeneous neural agents.
\begin{ack}
This work was supported by the SNS JU project 6G-GOALS under the EU Horizon Europe program, Grant Agreement No. 101139232, and by Huawei Technology France SASU under Grant No. Tg20250616041.
\end{ack}



\bibliographystyle{unsrt}
\bibliography{references}

\appendix
\newpage
 
\addcontentsline{toc}{section}{Appendix}
 
\part{Supplementary Material}
\mtcsetdepth{parttoc}{1}
{\hypersetup{hidelinks}
\parttoc}

\newpage

\section{Data Extraction and Organization}
\label{app:data_org}

\subsection{Computational Infrastructure and Latent Space Extraction}
\label{app:flop_cost}

Extracting the latent representations used in this work constitutes the most      computationally demanding phase of the pipeline. We encoded the embeddings of $1{,}697$ pretrained models drawn from the \texttt{timm} library across eight standard vision benchmarks: CIFAR-10, CIFAR-100, Tiny-ImageNet, ImageNet-1k, MNIST, Fashion-MNIST, CelebA, and Oxford Flowers. The extraction was carried out on a heterogeneous GPU cluster consisting of two NVIDIA RTX 3090 GPUs (24\,GB VRAM each) and two NVIDIA RTX 4090 GPUs (8\,GB and 24\,GB VRAM respectively).

The total computational cost of this phase scales as
\[
    \mathcal{C} \;=\; \sum_{m=1}^{M} \mathrm{FLOPs}(m) \times N,
\]
where $M = 1{,}697$ is the number of models, $\mathrm{FLOPs}(m)$ denotes the floating-point operation count of a single forward pass through model $m$, and $N = \sum_{d} |\mathcal{D}_d|$ is the total number of observations aggregated across all eight dataset benchmarks $\{\mathcal{D}_d\}$. Given the diversity
of architectures in \texttt{timm} this cost is substantial and non-trivial to reproduce.

Crucially, this one-time extraction cost need not be borne by future researchers. All precomputed latent spaces are made publicly available, removing the need to re-run any model inference. This democratises access to large-scale representation benchmarking in two distinct ways. First, it eliminates the need for high-end GPU infrastructure and days of computation: researchers can directly load the precomputed embeddings and focus on the alignment methods themselves, reducing the effective cost from $\mathcal{O}(M \cdot N)$ forward passes to a simple data download. Second, and more significantly, it opens access to the latent spaces of large-scale models that would otherwise be inaccessible to many research groups. Several models in the \texttt{timm} library require substantial GPU memory even at inference time, placing them out of reach for researchers without access to high-end hardware. By releasing their precomputed embeddings, we decouple the study of latent space alignment from the hardware requirements of the underlying models, allowing any researcher with modest computational resources to work with representations from state-of-the-art architectures.

\subsection{Benchmark Datasets}
\label{app:datasets}

This appendix details the benchmark datasets used to construct the \textsc{Semasia} collection. For each dataset, we describe the domain, the number of classes and samples, the native image resolution and color format, the data splits on which latent representations were extracted, and the structure of the label columns inherited from the source dataset and preserved in our Parquet files. A summary is reported in Table~\ref{tab:datasets}.

\paragraph{Label-handling convention.}
Across all \textsc{Semasia} datasets, label columns are inherited verbatim from the corresponding Hugging Face source repository, with the only modification being the addition of the \texttt{id} column that pairs every latent representation with its source example. Whenever the source repository exposes a single class label, we preserve it under the name \texttt{label} as a \texttt{ClassLabel} feature, mapping integer indices to human-readable class names through the metadata in the dataset card. Whenever the source repository exposes multi-attribute or fine/coarse label hierarchies (as in CelebA and CIFAR-100, respectively), we preserve every label column individually so that downstream users can choose the granularity most relevant for their analysis. The full list of label columns for each dataset, together with their data types and value ranges, is given in the per-dataset paragraphs below.

\textbf{MNIST}~\citep{LeCun1998} is a classical benchmark for handwritten digit recognition. It contains $70{,}000$ grayscale images of digits from $0$ to $9$, each of resolution $28\!\times\!28$ pixels, partitioned into $60{,}000$ training and $10{,}000$ test samples. We extract latent representations on both splits. The \texttt{semasia-mnist} dataset exposes a single label column, \texttt{label} (\texttt{int64}, range $0$--$9$), corresponding to the digit class.\\
\small \hflogo \, Original Dataset: \url{https://huggingface.co/datasets/ylecun/mnist}.
\normalsize

\textbf{Fashion-MNIST}~\citep{Xiao2017} is a drop-in replacement for MNIST designed to be more challenging while preserving the same format. It contains $70{,}000$ grayscale images at $28\!\times\!28$ resolution, evenly distributed across $10$ clothing-item classes (e.g., \textit{t-shirt}, \textit{trouser}, \textit{sneaker}), with the canonical $60{,}000/10{,}000$ train/test split. We extract latent representations on both splits. The \texttt{semasia-fashion-mnist} dataset provides a single label column, \texttt{label} (\texttt{int64}, range $0$--$9$), with class names available in the source dataset card.\\
\small \hflogo \, Original Dataset: \url{https://huggingface.co/datasets/zalando-datasets/fashion_mnist}.
\normalsize

\textbf{CIFAR-10}~\citep{Krizhevsky2009} consists of $60{,}000$ RGB natural images at $32\!\times\!32$ resolution, organized into $10$ mutually exclusive object categories (\textit{airplane}, \textit{automobile}, \textit{bird}, \textit{cat}, \textit{deer}, \textit{dog}, \textit{frog}, \textit{horse}, \textit{ship}, \textit{truck}) with $6{,}000$ images per class and the canonical $50{,}000/10{,}000$ train/test split. We extract latent representations on both splits. The \texttt{semasia-cifar10} dataset exposes a single label column, \texttt{label} (\texttt{int64}, range $0$--$9$).\\
\small \hflogo \, Original Dataset: \url{https://huggingface.co/datasets/uoft-cs/cifar10}.
\normalsize

\textbf{CIFAR-100}~\citep{Krizhevsky2009} extends CIFAR-10 to $100$ fine-grained classes, grouped into $20$ semantic superclasses. It contains $60{,}000$ RGB images at $32\!\times\!32$ resolution, with $600$ images per class and the standard $50{,}000/10{,}000$ train/test split. We extract latent representations on both splits. The \texttt{semasia-cifar100} dataset exposes two label columns, both inherited from the source: the fine label \texttt{fine\_label} (\texttt{int64}, range $0$--$99$) and the coarse label \texttt{coarse\_label} (\texttt{int64}, range $0$--$19$). Preserving both granularities makes this dataset particularly suitable for probing the hierarchical organization of learned representations.\\
\small \hflogo \, Original Dataset: \url{https://huggingface.co/datasets/uoft-cs/cifar100}.
\normalsize

The \textbf{Oxford 102 Flowers} dataset~\citep{Nilsback2008} is a fine-grained classification benchmark consisting of $8{,}189$ RGB images of flowers commonly found in the United Kingdom, distributed across $102$ categories with between $40$ and $258$ images per class. The dataset is split into $7{,}170$ training, and $1{,}020$ test images. We extract latent representations on all three splits. The \texttt{semasia-oxford-flowers} dataset exposes a single label column, \texttt{label} (\texttt{int64}, range $0$--$101$), with class names recoverable from the source dataset card.\\
\small \hflogo \, Original Dataset: \url{https://huggingface.co/datasets/nkirschi/oxford-flowers}.
\normalsize

\textbf{Tiny ImageNet}~\citep{Le2015} is a downsampled subset of ImageNet introduced for the Stanford CS231N course. It contains $200$ classes with $100{,}000$ training, and $10{,}000$ validation images per class, all rescaled to $64\!\times\!64$ resolution. We extract latent representations on the training and validation splits, since test labels are not publicly released. The \texttt{semasia-tiny-imagenet} dataset exposes a single label column, \texttt{label} (\texttt{int64}, range $0$--$199$), where each integer corresponds to a WordNet synset (\texttt{wnid}) preserved in the source repository for cross-referencing with the full ImageNet hierarchy.\\
\small \hflogo \, Original Dataset: \url{https://huggingface.co/datasets/zh-plus/tiny-imagenet}.
\normalsize

\textbf{ImageNet-1k}~\citep{Deng2009, Russakovsky2015} is the de facto standard large-scale benchmark for image classification. It comprises approximately $1.28$ million training images, $50{,}000$ validation, and $100{,}000$ test images, organized into $1{,}000$ object categories drawn from the WordNet hierarchy. Images are RGB and of variable resolution, typically resized and center-cropped to $224\!\times\!224$ during preprocessing. We extract latent representations on the validation split. The \texttt{semasia-imagenet1k} dataset exposes a single label column, \texttt{label} (\texttt{int64}, range $0$--$999$), with each integer corresponding to a WordNet synset whose human-readable description is provided in the source dataset card.\\
\small \hflogo \, Original Dataset: \url{https://huggingface.co/datasets/ILSVRC/imagenet-1k}.
\normalsize

\textbf{CelebFaces Attributes} (CelebA)~\citep{Liu2015} is a large-scale face attributes dataset containing $202{,}599$ celebrity face images from $10{,}177$ unique identities. We use the aligned-and-cropped version of the dataset and follow the official $162{,}770/19{,}867/19{,}962$ train/validation/test split, extracting latent representations on all three. CelebA is the only benchmark in \textsc{Semasia} with multi-label annotations: the \texttt{semasia-celeba} dataset exposes $40$ binary attribute columns, one per attribute (e.g., \texttt{Smiling}, \texttt{Eyeglasses}, \texttt{Young}, \texttt{Male}, \texttt{Wearing\_Hat}, each \texttt{int64} with values in $\{0, 1\}$), preserving the exact column naming of the source repository. The $5$ landmark coordinates available in the original CelebA release are not retained, since they are not relevant to the classification analyses targeted by \textsc{Semasia}; users requiring landmark information can join \textsc{Semasia} rows with the source repository through the \texttt{id} column. This rich multi-attribute structure enables fine-grained probing of attribute disentanglement in latent representations.\\
\small \hflogo \, Original Dataset: \url{https://huggingface.co/datasets/flwrlabs/celeba}.
\normalsize

\subsection{Model Registry}
\label{app:registry}

\begin{figure}[t]
    \centering
    \includegraphics[width=\textwidth]{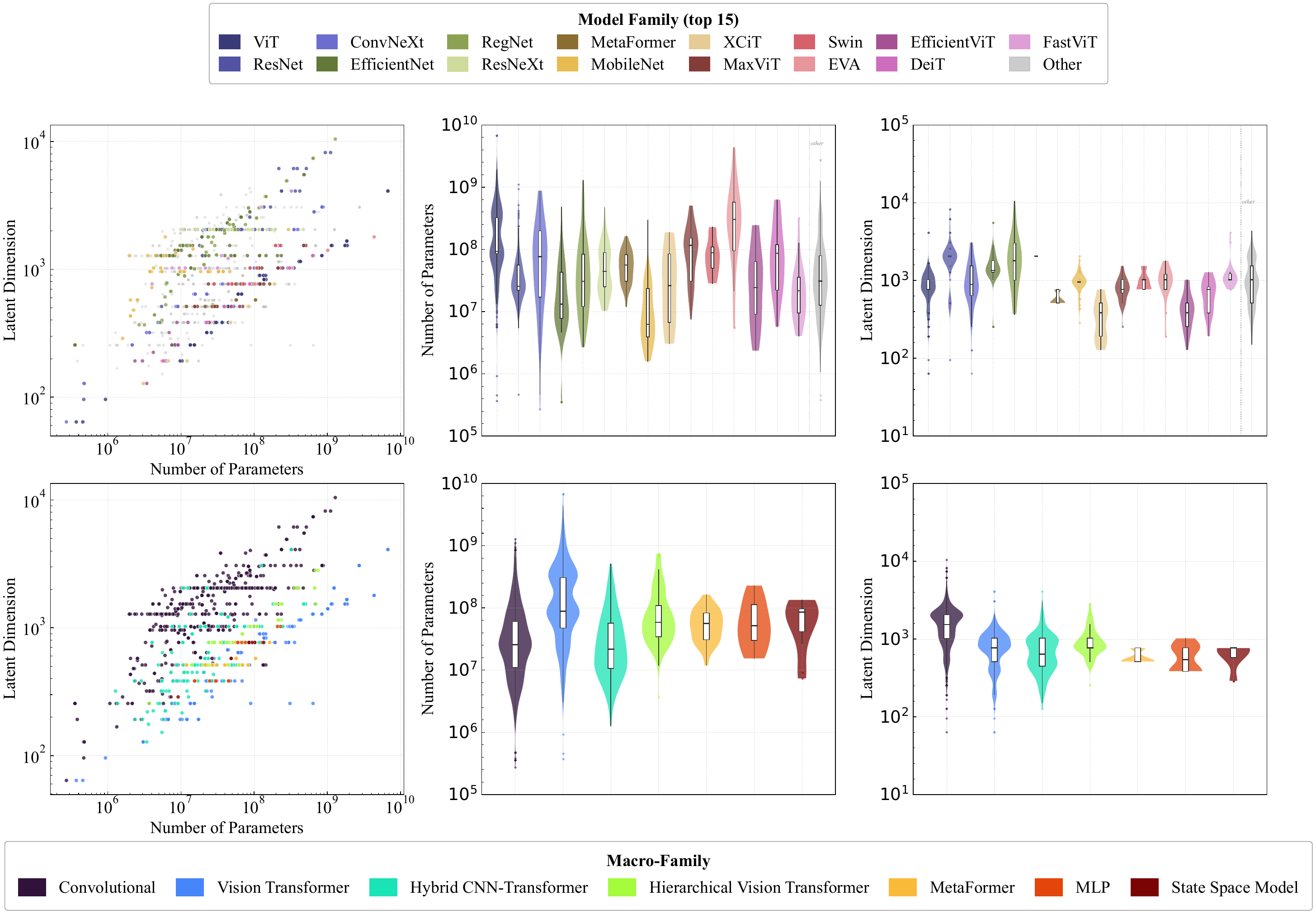}
    \caption{Exploratory analysis of the \textsc{Semasia} model registry. \textbf{Left:} joint distribution of the number of trainable parameters and the latent space dimensionality, shown on a log--log scale. \textbf{Center:} marginal distribution of the number of parameters across models. \textbf{Right:} marginal distribution of the latent space dimensionality. The bottom row aggregates models by architectural macro-family, while the top row provides a fine-grained breakdown by model family, showing the fifteen most populated families and grouping the remaining ones under \textit{Others}.}
    \label{fig:eda}
\end{figure}

The \textsc{Semasia} model registry is a tabular companion dataset collecting metadata for every \texttt{timm} architecture used to extract latent representations. Its design follows the \texttt{timm} identifier convention: \emph{everything before the dot encodes the architecture, everything after encodes how and where the model was pretrained.} The registry mirrors this split, organizing columns into six thematic groups: identity, architecture, head and attention, variant flags, pretraining, and capacity.

\paragraph{Identity.}
The \texttt{model\_name} field stores the full \texttt{timm} identifier (e.g., \texttt{vit\_large\_patch14\_clip\_224.openai\_ft\_in1k}) and serves as the primary key linking the registry to the per-row \texttt{model\_name} column in each semantic dataset, so any joint analysis reduces to a join on this column.

\paragraph{Architecture.}
Parsed from the pre-dot portion of the identifier, this group decomposes architectural choices into nine fields. \texttt{family} captures the high-level class (\textit{ViT}, \textit{ConvNeXt}, \textit{ResNet}, \textit{Swin}, \textit{DeiT}, \textit{EfficientNet}, etc.), optionally refined by \texttt{model\_version} (\textit{v2}, \textit{v3}, \textit{SE}, \textit{NF}, \textit{CSP}, \dots) for multi-generation families. Capacity is described by three complementary fields: a human-readable \texttt{size} label (\textit{Tiny}--\textit{Giant}, or EfficientNet B$0$--B$8$), a numeric \texttt{depth\_code} for depth-scaled families (e.g., \texttt{50}/\texttt{101} for ResNets, \texttt{f0}--\texttt{f6} for NFNets), and a \texttt{width\_code} for channel/group-width multipliers (e.g., \texttt{32x4d} for ResNeXt, \texttt{w44} for HRNet, \texttt{075} for MobileNet). Geometric parameters are captured by \texttt{patch\_size} (ViT only, \textit{null} for CNNs), \texttt{input\_resolution} (the native architectural resolution, distinct from pretraining/fine-tuning resolution), \texttt{window\_size} (Swin/MaxViT/CoAtNet local-attention window, \textit{null} for global attention), and \texttt{stride\_code} (a stage/stride code for families such as CaiT, ConvFormer, and ShViT).

\paragraph{Head and attention.}
This group captures pooling and attention choices. \texttt{head\_type} reports the pooling strategy: \texttt{GAP} (global average pooling), \texttt{CLS} (class-token), or \texttt{CLS+GAP} (hybrid). \texttt{num\_registers} records DINOv2-style register tokens (typically $1$ or $4$). \texttt{positional\_encoding} flags non-default schemes (\texttt{RoPE}, \texttt{RelPos}, \texttt{SinCos}, \texttt{APE}), with \textit{null} for the default learned absolute encoding. \texttt{activation} flags non-default activations, notably \texttt{QuickGELU} for OpenAI-style CLIP models (\textit{null} for standard GELU/ReLU). Finally, \texttt{pe\_scope} records the ViT-PE scope tag (\textit{Lang}, \textit{Core}, \textit{Spatial}).

\paragraph{Variant flags.}
A set of booleans records architectural and training variants. Training-regime flags: \texttt{is\_distilled} (e.g., DeiT with a RegNet teacher), \texttt{is\_pruned} (structured pruning), and \texttt{is\_legacy} (older models under the \texttt{legacy\_} prefix). \texttt{is\_gap} and \texttt{uses\_quickgelu} are shorthands for \texttt{head\_type == "GAP"} and \texttt{activation == "QuickGELU"}, included to simplify filtering. Family-specific micro-architectural flags: \texttt{uses\_rmlp} (MaxViT/CoAtNet with MLP Log-CPB relative position bias for resolution-generalizable attention); \texttt{uses\_rw} (\texttt{timm} re-implementations tuned for PyTorch eager-mode efficiency); \texttt{uses\_cr} (SwinV2 cross-resolution variants); \texttt{uses\_ns} (SwinV2-CR norm-per-stage variants applying LayerNorm at every stage); \texttt{uses\_abswin} (Hiera with absolute window position embeddings); \texttt{uses\_ts} (BYOBNet with a tiered three-layer convolutional stem); and \texttt{uses\_aa} (anti-aliased downsampling).

\paragraph{Pretraining.}
Parsed from the post-dot portion of the identifier, this group describes how and where the checkpoint was produced. The raw configuration string is preserved under \texttt{pretrain\_config} for reference and debugging. \texttt{pretrain\_org} records the training organization (e.g., \textit{Meta}, \textit{Apple}, \textit{OpenAI}, \textit{Google}, \emph{timm SBB recipe}). The corpus is described by \texttt{pretrain\_dataset} (\textit{ImageNet-1K/21K/22K}, \textit{LAION-2B}, \textit{WebLI}, \textit{LVD-142M}, \dots) and, when ambiguous, by \texttt{pretrain\_dataset\_size} (\textit{400M}, \textit{s39b}, \textit{2.1T}). \texttt{pretrain\_method} captures the objective (\textit{CLIP}, \textit{SigLIP}, \textit{MAE}, \textit{DINO}, \textit{DINOv2}, \textit{MIM}, \textit{FCMAE}, \textit{AugReg}, \dots; \textit{null} for standard supervised training). \texttt{pretrain\_ft} records subsequent fine-tuning (typically \textit{ImageNet-1K}, occasionally \textit{ImageNet-22K} or \textit{ImageNet-12K}). Resolution differences are split across \texttt{pretrain\_resolution} and \texttt{pretrain\_ft\_resolution}. Compute budget is encoded either in \texttt{pretrain\_epochs} (e.g., \texttt{e200} $\to 200$) or, for token-budget recipes such as SAM2, in \texttt{pretrain\_tokens} (e.g., \texttt{2pt1} $\to 2.1$T tokens). \texttt{pretrain\_aug} tracks augmentation (\textit{AugReg}, \textit{AdvProp}, \textit{NoisyStudent}, \textit{AutoAugment}, \textit{RandAugment}). Finally, \texttt{pretrain\_i18n} flags SigLIP models pretrained on multilingual WebLI ($109$ languages).

\paragraph{Model capacity.}
Two derived fields are obtained by instantiating each model in \texttt{timm} (without downloading weights, when feasible). \texttt{num\_parameters} reports total trainable parameters, and \texttt{latent\_dim} reports the output dimensionality of \texttt{forward\_features()}, i.e., the dimensionality of the vector stored in the \texttt{embedding} column of the corresponding \textsc{Semasia} dataset. Together they enable scaling analyses and comparisons of representational capacity across the registry.

The model registry metadata are leveraged to analyze mismatches across models from different architectural families, as well as differing pre-training and fine-tuning regimes, highlighting the analytical potential of SEMASIA as the first large-scale collection of latent spaces enabling controlled statistical inference over representations. To illustrate the diversity of the collection, Figure~\ref{fig:eda} presents a preliminary exploratory analysis of two key \textbf{model capacity} descriptors: the number of trainable parameters and the dimensionality of the latent space at the selected extraction layer. The left panels reveal a clear power-law relationship between these quantities, indicating sublinear scaling of latent dimensionality with model size: increases in parameter count correspond to proportionally smaller increases in representation width on a logarithmic scale. However, this aggregate trend conceals substantial heterogeneity across architectural families. As shown in the central and right panels, transformer-based and convolutional models occupy distinct regions of the parameter–latent dimension space, each exhibiting characteristic latent widths and scaling behaviors.

\section{Semantic Bottleneck of Simple Neural Models}
\label{app:cutting-point}

\begin{figure}[t]
\centering
\includegraphics[width=\linewidth]{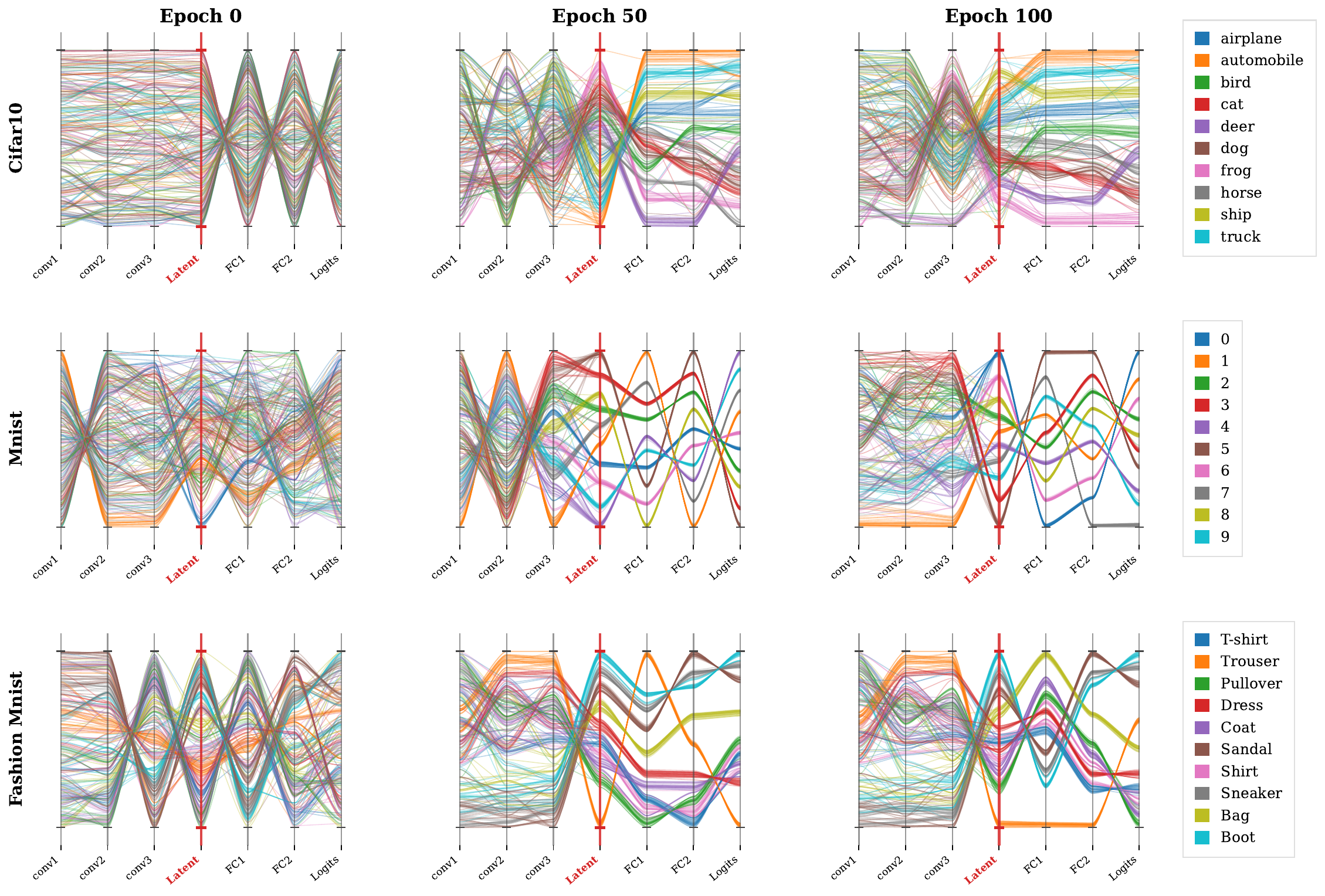}
\caption{Evolution of the representation space learned by the convolutional classifier across three datasets (MNIST, Fashion-MNIST, CIFAR-10) and three training checkpoints. At each layer, 1D t-SNE is applied to the test-set representations, with colors denoting class labels. The semantic bottleneck emerges progressively within the encoder, while the decoder introduces no further semantic reorganization.}
\label{fig:semantic_bottleneck_classifier}
\end{figure}

\begin{figure}[t]
\centering
\includegraphics[width=\linewidth]{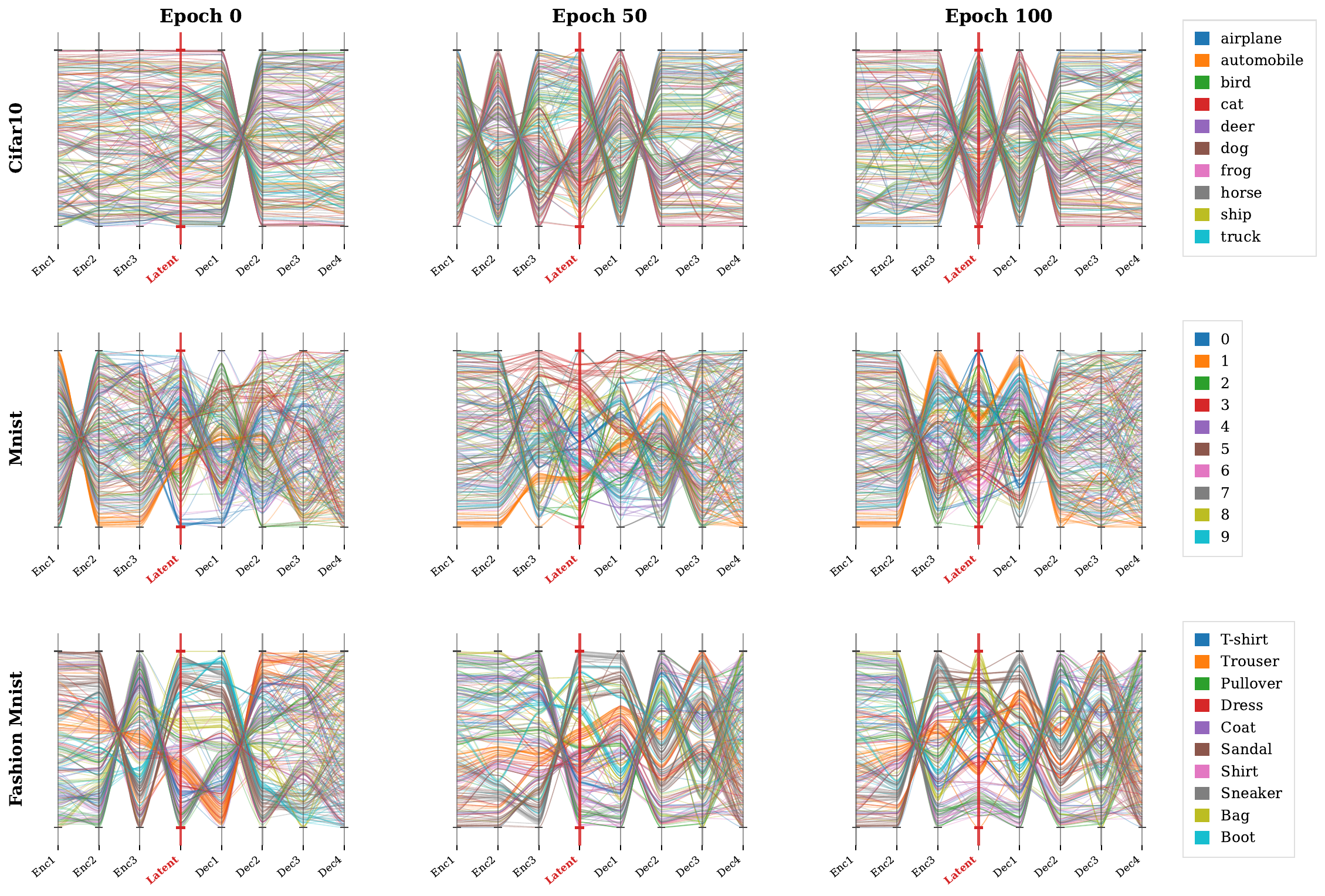}
\caption{Evolution of the representation space learned by the convolutional autoencoder across three datasets (MNIST, Fashion-MNIST, CIFAR-10) and three training checkpoints. At each layer, 1D t-SNE is applied to the test-set representations, with colors denoting class labels. Unlike the classification setting, the reconstruction objective does not induce a semantically organized bottleneck, highlighting the critical role of the downstream task in shaping representation geometry.}
\label{fig:semantic_bottleneck_autoencoder}
\end{figure}

We present two controlled experiments designed to empirically validate the central claims of our framework. The first studies the emergence of the semantic bottleneck in a classification setting; the second examines how the nature of the downstream task shapes the geometry of the learned representations. Taken together, these experiments demonstrate that the semantic bottleneck is not an intrinsic property of the architecture alone, but is actively shaped by the objective the model is trained to optimize.
\subsection{Semantic Bottleneck in Classification}
\paragraph{Setup.} We consider image classification on three standard benchmarks: MNIST, Fashion-MNIST, and CIFAR-10. The model follows an encoder-decoder architecture. The encoder consists of four convolutional layers with channel widths doubling at each stage, each followed by batch normalization, ReLU activation, max pooling, and dropout. The decoder is a two-layer fully connected network with ReLU activations, culminating in a linear projection to class logits.
\paragraph{Emergence of the semantic bottleneck.} Figure~\ref{fig:semantic_bottleneck_classifier} traces the evolution of the representation space across layers and training stages, evaluated on held-out test data. At each layer, we apply 1D t-SNE to visualize the geometry of the learned representations. At initialization, the network produces unstructured embeddings with no discernible semantic organization. As training progresses, a clear bottleneck structure emerges in the latent space: representations become increasingly clustered, reflecting the semantic content of the input.
This structure manifests at two complementary levels of granularity. At the \emph{micro} level, class-conditional inputs form compact, well-separated clusters, providing the geometric basis for downstream classification. At the \emph{macro} level, semantically related classes coalesce into broader super-clusters without explicit supervision: in CIFAR-10, animal and vehicle categories form two clearly delineated groups, while in Fashion-MNIST, footwear and clothing are similarly distinguished. This unsupervised emergence of hierarchical semantic organization suggests that the encoder discovers latent structure that generalizes beyond the classification objective itself.
\paragraph{Role of the decoder.} Representations extracted from the classifier head introduce no additional semantic structure beyond what is already present at the bottleneck: they constitute permutations of the clusters formed in the encoder's latent space. This finding has two important implications. First, it empirically validates our architectural choice of locating the semantic bottleneck at the encoder-decoder interface. Second, it provides evidence that the locus of semantic compression is the encoder, not the classifier, lending empirical support to a semantic theory of learned representations.

\subsection{The Role of the Downstream Task: Reconstruction}
\paragraph{Setup.} To isolate the effect of the downstream task on the geometry of learned representations, we train a convolutional autoencoder on the same three benchmarks. The encoder follows the same architecture as above. The decoder mirrors it with transposed convolutions, reconstructing the input from the bottleneck representation. Unlike the classification setting, no explicit semantic signal is provided during training: the sole objective is pixel-level reconstruction.
\paragraph{Task-dependent geometry.} Figure~\ref{fig:semantic_bottleneck_autoencoder} reveals a striking contrast with the classification case. Under the reconstruction objective, the latent space lacks the well-defined cluster structure observed previously. We interpret this as evidence that the downstream task actively guides the geometry of the representation space: without a semantic objective, the encoder is not incentivized to organize representations according to class identity or semantic similarity, but merely to preserve enough information for faithful reconstruction.
While the latent spaces for MNIST and Fashion-MNIST exhibit a modest degree of grouping — likely attributable to perceptual similarity between inputs within the same category — this structure is considerably weaker and less consistent than in the classification setting. Crucially, the decoder plays an entirely different role here: rather than preserving the semantic geometry of the bottleneck, it must invert it, mapping compressed representations back to pixel space and thereby dissolving any aggregated semantic structure present at the latent level.
\paragraph{Task as a geometric prior.} Jointly, these two experiments support a central tenet of our framework: the semantic bottleneck is not an emergent property of depth or nonlinearity alone, but is induced by the interplay between architecture and objective. The downstream task acts as a geometric prior on the representation space, and it is the classification objective — with its requirement to map semantically distinct inputs to distinct outputs — that drives the formation of a structured, semantically organized bottleneck.
\section{Concept Clustering}
\label{app:concept-clustering}

We extend the analysis of Section~\ref{sec:lse} by examining concept clustering in the latent space across different granularities. The focus is on how representations distribute along principal directions of the embedding, in analogy with a long line of work that interprets dominant axes of a learned representation as carriers of semantic content.

\paragraph{Latent semantic analysis in NLP.}
The earliest and most influential instance of this idea is Latent Semantic Analysis (LSA)~\citep{deerwester1990indexing,landauer1997solution}, where a truncated singular value decomposition (SVD) of a term–document matrix recovers latent semantic factors aligned with the leading singular directions. This linear-algebraic view of meaning was later refined by probabilistic latent variable models such as probabilistic LSA~\citep{hofmann2001unsupervised} and Latent Dirichlet Allocation~\citep{blei2003latent}, and extended to distributed word representations whose principal directions encode syntactic and semantic regularities~\citep{mikolov2013distributed,pennington2014glove,levy2014neural}. More recent analyses of contextual embeddings produced by transformer language models show that a small number of principal components captures most of the linguistic variance and isolates interpretable factors such as part-of-speech, sentiment, or topic~\citep{ethayarajh2019contextual,reif2019visualizing,cai2021isotropy}. Across these works, the recurring pattern is that linear projections onto leading components of an SVD- or PCA-style decomposition expose structure that is otherwise entangled in the raw representation.

\begin{figure}[t]
    \centering
    \includegraphics[width=0.7\textwidth,trim={2cm 3cm 4.55cm 1cm}, clip]{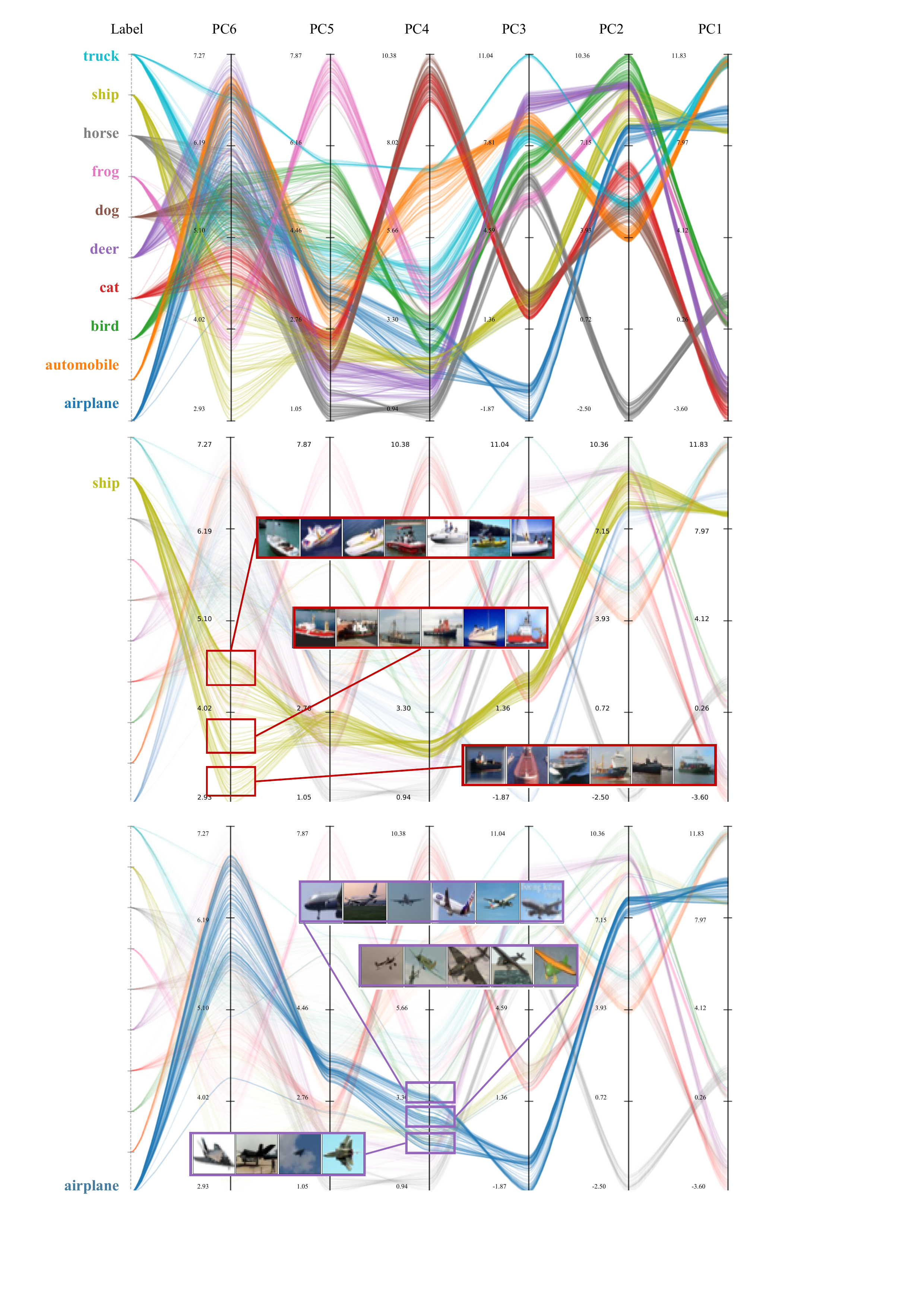}
\caption{Parallel coordinates of the t-SNE projection of the latent space of \texttt{aimv2\_1b\_patch14\_224.apple\_pt} on CIFAR-10. Lines denote samples colored by class. Top: full dataset. Middle/bottom: zoom on \emph{ship} and \emph{airplane}, highlighting intra-class structure.}
    \label{fig:cifar10}
\end{figure}

\paragraph{Principal directions in visual representations.}
A parallel line of work in computer vision interprets principal axes of learned features as visual concept directions. Classical results on Eigenfaces~\citep{turk1991face} already showed that PCA on aligned face images yields components that correspond to coarse identity and illumination factors. In modern deep representations, PCA and related linear probes on convolutional and transformer features have been used to identify part- and object-level concepts~\citep{zhang2018unreasonable,kornblith2019better}, while the authors of \citep{voynov2020unsupervised} and \citep{harkonen2020ganspace} show that the leading principal directions of GAN latent and feature spaces correspond to interpretable image edits such as zoom, rotation, age, or lighting. Concept-activation methods such as TCAV~\citep{kim2018interpretability} and concept-bottleneck models~\citep{koh2020concept} formalize the idea that human-interpretable concepts live along low-dimensional linear subspaces of deep representations.
\begin{figure}[t]
    \centering
    \includegraphics[width=\textwidth,trim={0 18cm 3.55cm 1cm}, clip]{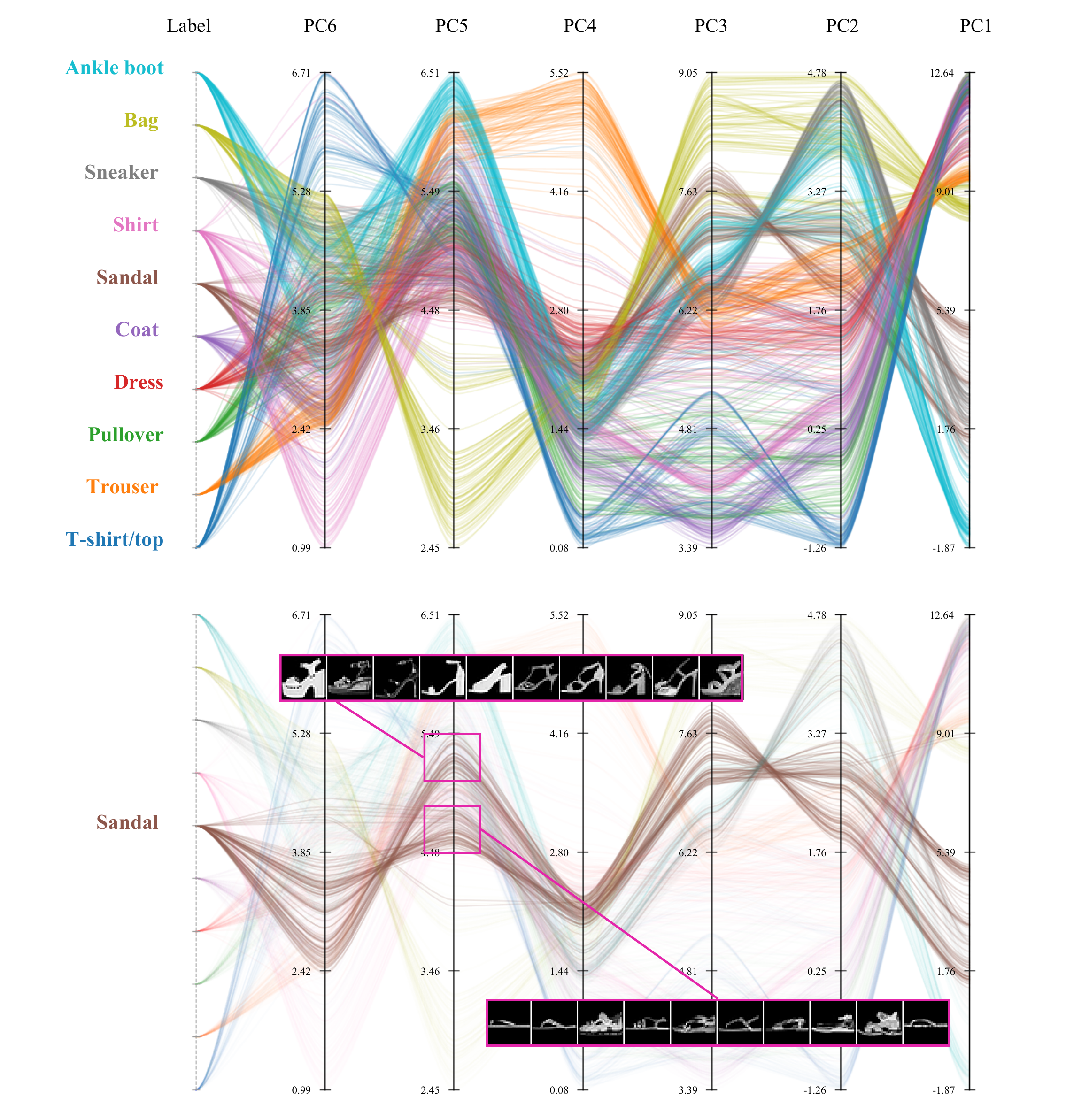}
\caption{Parallel coordinates of the t-SNE projection of the latent space on Fashion-MNIST. The embedding separates garment categories and reveals intra-class structure (e.g., sandals split by structural attributes such as heels).}
    \label{fig:fashion_mnist}
\end{figure}

\paragraph{Nonlinear projections and visualization.}
Beyond linear factorizations, nonlinear dimensionality reduction methods such as t-SNE~\citep{van2008visualizing} and UMAP~\citep{mcinnes2018umap} preserve local neighborhood structure. Although the coordinates produced by t-SNE and UMAP have no closed-form interpretation as eigenvectors of an underlying operator, they are routinely treated as semantic axes in their own right: in single-cell biology, individual UMAP coordinates are interpreted as developmental or phenotypic gradients~\citep{becht2019umap,kobak2019art}, and in representation analysis, they are inspected directly to recover class structure and intra-class variation. Visualizing such embeddings via parallel coordinates~\citep{inselberg1985plane} makes the per-axis distribution of samples explicit and allows class-conditional patterns along each UMAP dimension to be read off directly. This is the strategy adopted below.

\begin{figure}[t]
    \centering
    \includegraphics[width=\textwidth,trim={2cm 27cm 0.5cm 4cm}, clip]{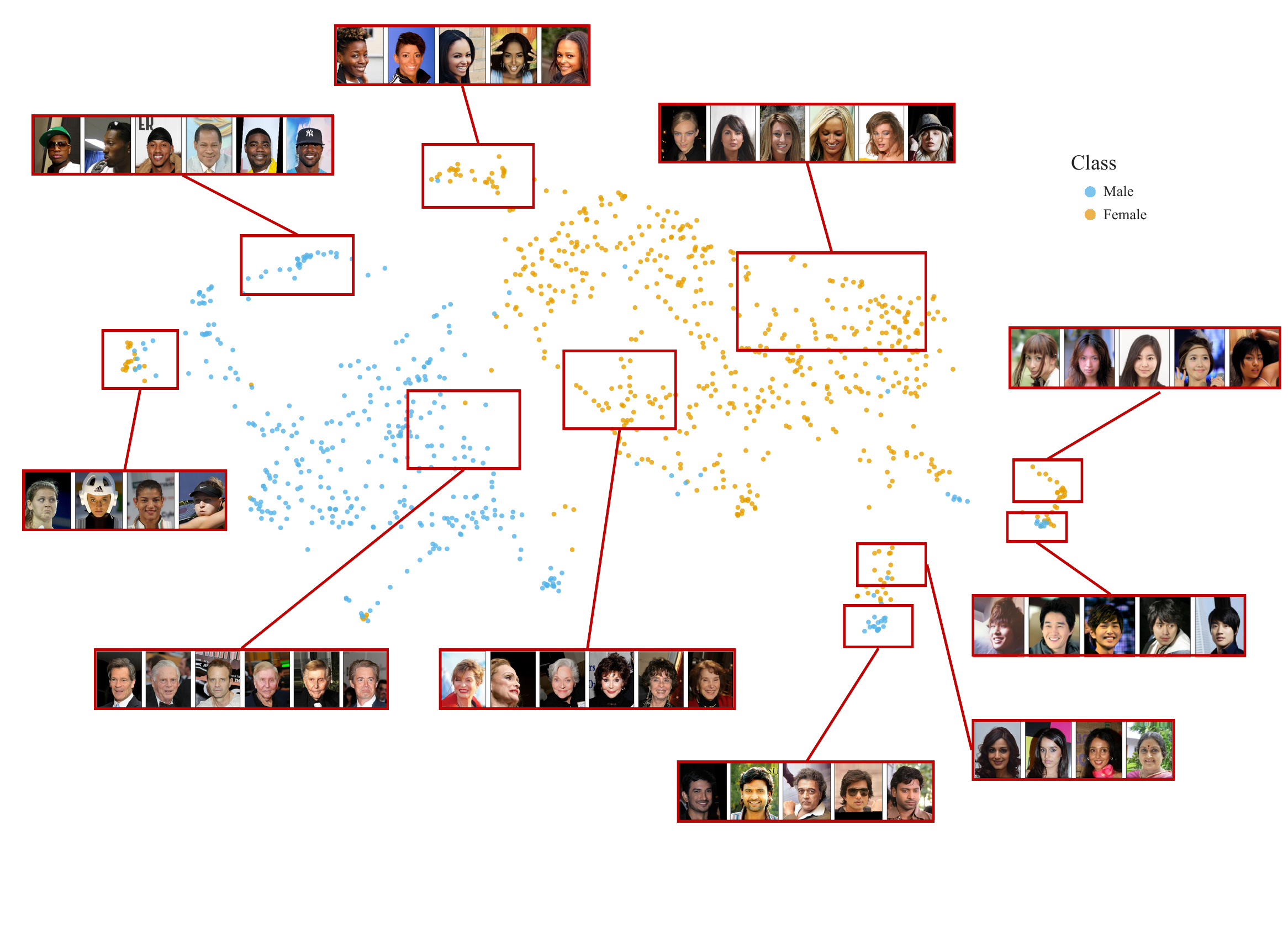}
\caption{2D UMAP projection of latent representations for CelebA. Points correspond to images. The embedding shows structured clustering driven by multiple facial attributes (e.g., gender, age, and other visual traits).}
    \label{fig:celeba}
\end{figure}

\paragraph{Empirical analysis on \textsc{Semasia}.}
We analyze the \texttt{aimv2\_1b\_patch14\_224.apple\_pt} latent space on multiple \textsc{Semasia} benchmarks using UMAP followed by parallel coordinate visualizations. Clear class-level clustering emerges, together with interpretable intra-class structure, mirroring the linear-semantic pattern observed in LSA and Eigenfaces but at the scale of modern self-supervised encoders.

On Fashion-MNIST (Figure~\ref{fig:fashion_mnist}), the embedding captures fine-grained visual attributes. For example, \emph{sandals} separate along a principal direction according to the presence of heels, recovering a structural attribute that was never explicitly supervised.

On CIFAR-10 (Figure~\ref{fig:cifar10}), samples cluster by class across UMAP dimensions, and individual components encode meaningful semantic variations. PC6 organizes the \emph{ship} class along a size continuum, from small private boats to large vessels such as container ships and tankers. For \emph{airplanes}, PC4 separates subcategories such as commercial aircraft, vintage planes, and fighter jets.

Figure~\ref{fig:celeba} shows a 2D UMAP projection for CelebA. The embedding exhibits structured clustering across multiple facial attributes, including gender, age, and finer visual traits such as hairstyle. Nearby regions correspond to visually similar subjects, while distinct groups remain well separated, in line with the Eigenfaces tradition~\citep{turk1991face} and with concept-bottleneck analyses of face representations~\citep{koh2020concept}.

These patterns reflect correlations in the learned visual representations induced by the syntactic relations among input examples from the dataset, without implying any normative interpretation of sensitive attributes. Taken together, they support the broader hypothesis, recurrent across NLP and vision, that semantically meaningful structure in learned representations is concentrated along a small number of dominant directions and is well exposed by combining linear factorizations with neighborhood-preserving projections.

\section{Semantic Mismatch}
\label{app:mismatch}

In this section, we examine the differences between the bases used to represent the latent spaces of pairs of models. As discussed in Section \ref{sec:lse} and Appendix \ref{app:concept-clustering}, these bases capture the principal directions along which latent representations are distributed and clusterized, serving as a proxy for latent features.

In addition, we characterize semantic mismatch through the analysis of concepts, defined as prototypical representations (centroids) within the latent space. Prototypes provide a natural entry point for probing the semantics encoded at different levels of granularity. By matching these prototypes across models, we can identify where their representations align and where they diverge in their internal organization of the data.

\subsection{Comparing Latent Bases}
\label{app:latent_bases}

\begin{figure}[t]
    \centering
    \includegraphics[width=\textwidth,trim={0 6cm 0 0}, clip]{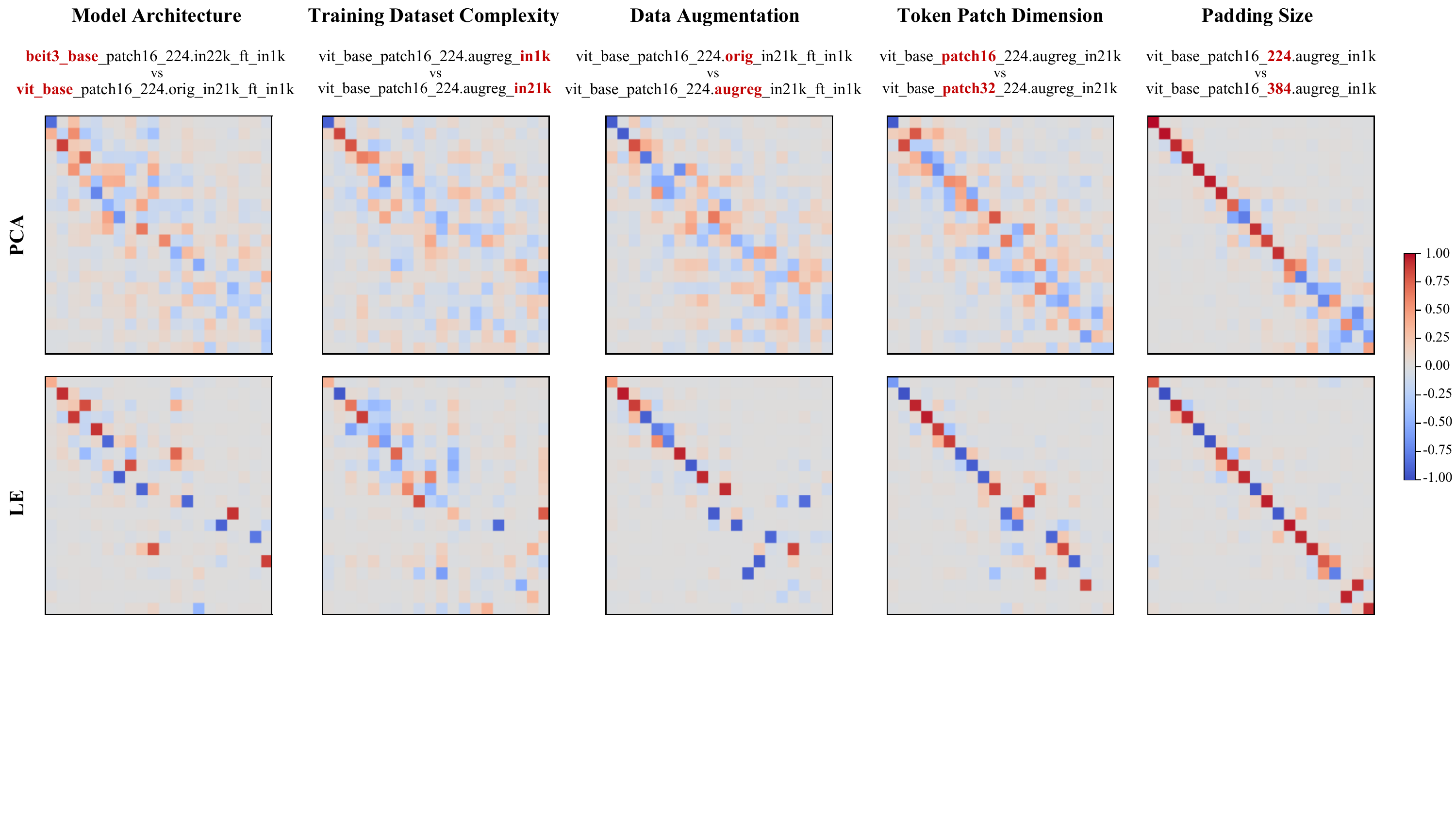}
    \caption{Cross-correlation heatmaps between basis vectors extracted from pairs of CIFAR-10 latent spaces~\citep{ovsjanikov2012functional}. Each column corresponds to a model pair differing along a single controlled source of heterogeneity: (i) architecture; (ii) pretraining data scale; (iii) data augmentation; (iv) tokenization patch size; and (v) padding size. Rows compare PCA (top) and Laplacian eigenmaps (bottom), truncated to the first twenty components.}
    \label{fig:pca}
\end{figure}

A natural way to compare two latent spaces is to inspect the bases that summarize their geometry. For each model, we compute two complementary bases from its CIFAR-10 latent point cloud: the first $20$ principal components and the first $20$ Laplacian eigenmaps~\citep{belkin2003laplacian} of a $k$-NN graph with $k=10$.

Figure~\ref{fig:pca} reports this comparison across five model pairs, each isolating a single source of heterogeneity. Two patterns emerge. First, basis mismatch tracks the magnitude of the perturbation, decreasing from architectural changes to padding, with patch size inducing a stronger effect than padding due to its impact on tokenization granularity~\citep{grimaldi2025learning}. Second, Laplacian eigenmaps are more robust than PCA, exhibiting structured cross-correlation patterns even when PCA bases appear unstructured. By encoding intrinsic neighborhood relations rather than ambient linear geometry, the $k$-NN graph captures a topological signature that is largely preserved under model-level perturbations. A quantitative analysis of these effects is provided in Appendix~\ref{app:stat_regression}.

\subsection{Concepts Correspondence}
\label{app:compare_concepts}

\begin{figure}[t]
    \centering
    \includegraphics[width=\textwidth,trim={0 14cm 12cm 0}, clip]{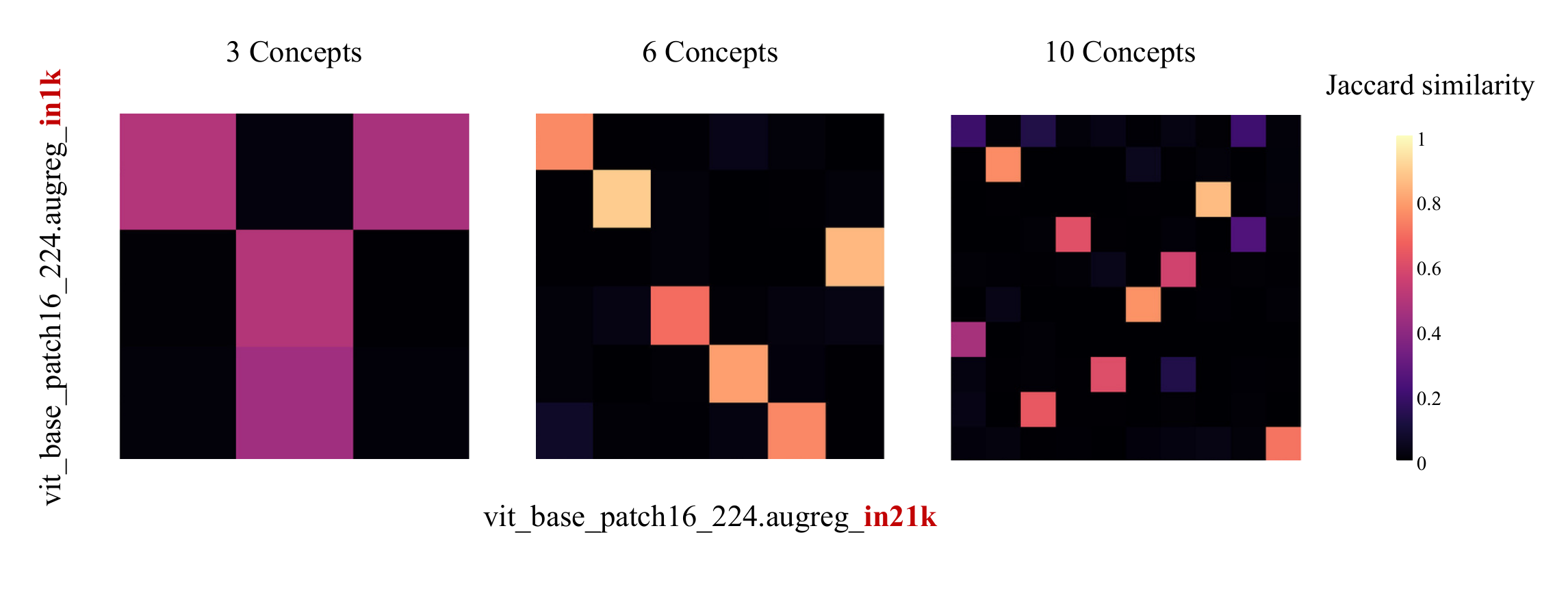}
    \caption{Jaccard similarity heatmaps between prototypical anchors extracted from two ViT models differing in pretraining data complexity. From left to right, $3$, $6$, and $10$ anchors are extracted per model.}
    \label{fig:proto}
\end{figure}

Basis mismatch alone cannot distinguish genuine differences in expressivity from reparameterizations of semantically equivalent representations~\citep{javidnia2026gauge}. We therefore turn to \emph{prototypical anchors}, defined as centroids of the latent point cloud, which provide a concept-level description of the representation.

Figure~\ref{fig:proto} reports cross-model concept similarity matrices for a representative model pair, varying the number of prototypes extracted according to Algorithm \ref{alg:proto_alg}. Compared to basis-level analyses, these matrices are significantly sparser, indicating that prototypical anchors capture a more localized and discriminative semantic structure. For a small number of prototypes, we observe a near one-to-one correspondence. As the number increases, correspondences become many-to-one, reflecting differences in semantic granularity between models.

\begin{algorithm}[t]
\caption{Prototypical Anchors}
\label{alg:proto_alg}
\begin{algorithmic}[1]
\STATE \textbf{Require:} a dataset $\mathcal{D}$, a desired number of clusters $\nclusters$ \textbf{or} an achors matrix $\mathcal{A}$, a number of samples $\nsamples$, a neural encoder $E$, and a complex compression mapping $\psi$.
\STATE \textbf{Return:} Index set $\mathcal{A}$ and prototypical anchor matrix $\mathbf{P}$.
\IF{$\mathcal{A}$ is not provided}
    \STATE $\mathcal{X} \leftarrow E(\mathcal{D})$.
    \STATE $\{\cluster_1, \dots, \cluster_\nclusters\} \leftarrow$ apply a clustering algorithm with $\nclusters$ clusters to $\mathcal{X}$ such that $\bigcup_{i=1}^{\nclusters} \cluster_i = \mathcal{X}$.
    \STATE $\mathcal{A} = \{\mathcal{A}_1, \dots, \mathcal{A}_\nclusters\} \leftarrow$ for each cluster $\cluster_i$, randomly sample $\nsamples$ indices to form $\mathcal{A}_i$.
\ENDIF
\STATE Compute the prototypical anchors matrix as \\
$\mathbf{P} = \{\mathbf{p}_1, \dots, \mathbf{p}_\nclusters\}$, where each prototype is computed as:
$$
\mathbf{p}_i = \frac{1}{\nsamples} \sum\nolimits_{\alpha \in \mathcal{A}_i} \psi(\mathcal{X}_\alpha).
$$
\STATE \textbf{return} $\mathcal{A}$ and $\mathbf{P}$.
\end{algorithmic}
\end{algorithm}

Given the $k \times k$ Jaccard similarity matrix $J$ (shown in Fig. \ref{fig:proto}), where each entry
\begin{equation}
    J_{ij} = \frac{|C_i^{\mathcal{A}} \cap C_j^{\mathcal{B}}|}{|C_i^{\mathcal{A}} \cup C_j^{\mathcal{B}}|}
\end{equation}
measures the overlap between prototype $i$ of model $\mathcal{A}$ and prototype $j$ of model $\mathcal{B}$ in terms of the data points they attract, we then consider three methods to identify correspondences between the two sets of prototypes.

\textbf{Hungarian matching}. We solve the Linear Sum Assignment Problem (LSAP) on $-J$ using the Hungarian algorithm \cite{kuhn1955hungarian}, finding the permutation $\pi^*$ that maximizes the total matched Jaccard similarity:
\begin{equation}
    \pi^* = \arg\max_{\pi} \sum_{i=1}^{k} J_{i,\pi(i)}.
\end{equation}
This yields a strict one-to-one correspondence between prototypes. The mean Jaccard similarity under this assignment,
\begin{equation}
    \mathcal{S}_{\mathrm{Hung}} = \frac{1}{k}\sum_{i=1}^{k} J_{i,\pi^*(i)},
\end{equation}

\textbf{Injected matching}. Each sample $x_i \in \mathcal{X}$ is assigned to the same cluster index $j \in \{1, \dots, k\}$ in both models, bypassing $\mathcal{B}$'s own clustering entirely. This constructs $k$ perfectly matched groups in which both models share the same partition of the data. The injected method allows us to isolate the contribution of $\mathcal{B}$'s learned geometry: discrepancies between the injected and independently clustered results reveal where the two encoders impose different partitions on the same data.

Formally, let $\mathcal{A} : \mathcal{X} \to \{1, \dots, k\}$ denote the clustering function of model $\mathcal{A}$, which assigns each sample $x_i \in \mathcal{X}$ to a cluster index $c_i = \mathcal{A}(x_i)$. 
In the injected matching scheme \cite{fiorellino2025frame}, the same assignment is forced onto model $\mathcal{B}$:
\begin{equation}
    c_i^{\mathcal{A}} = c_i^{\mathcal{B}} := \mathcal{A}(x_i), 
    \quad \forall\, x_i \in \mathcal{X},
\end{equation}
so that the $k$ groups $\{G_j\}_{j=1}^{k}$ are defined by a single shared partition:
\begin{equation}
    G_j = \bigl\{\, x_i \in \mathcal{X} \;\mid\; \mathcal{A}(x_i) = j \,\bigr\},
    \quad j = 1, \dots, k.
\end{equation}
This contrasts with the independent clustering scheme, in which $\mathcal{B}$ produces its own assignment $\hat{c}_i^{\mathcal{B}} = \mathcal{B}(x_i)$, potentially yielding a different partition of $\mathcal{X}$.

\textbf{Spectral matching}. We embed the $2k$ prototype nodes jointly using the eigenvectors of the normalized graph Laplacian of the symmetric adjacency matrix
\begin{equation}
    A = \begin{pmatrix} 0 & J \\ J^\top & 0 \end{pmatrix} 
    \in \mathbb{R}^{2k \times 2k}.
\end{equation}
The number of groups $k_{\mathrm{est}}$ is estimated automatically via the spectral gap:
\begin{equation}
    k_{\mathrm{est}} = \arg\max_{\ell}\,(\lambda_{\ell+1} - \lambda_{\ell}) + 1,
\end{equation}

where $\{\lambda_\ell\}$ are the sorted eigenvalues of the normalized Laplacian. The rows of the leading $k_{\mathrm{est}}$ eigenvectors, normalized to unit norm, are then clustered with $k$-Means, grouping nodes that are jointly well-connected in the Jaccard graph. Unlike Hungarian matching, spectral matching can recover many-to-many correspondences and does not require a similarity threshold, making it suitable for detecting merge/split phenomena where a single semantic region in one model is distributed across multiple clusters in the other.

We qualitatively evaluate the three matching methods 
by visualizing the prototype correspondences identified between 
\texttt{vit\_base\_patch16\_224.augreg\_in1k} and 
\texttt{vit\_base\_patch16\_224.augreg\_in21k} on two datasets of increasing 
semantic complexity.

\begin{figure}[h]
    \centering
    \includegraphics[width=\textwidth]{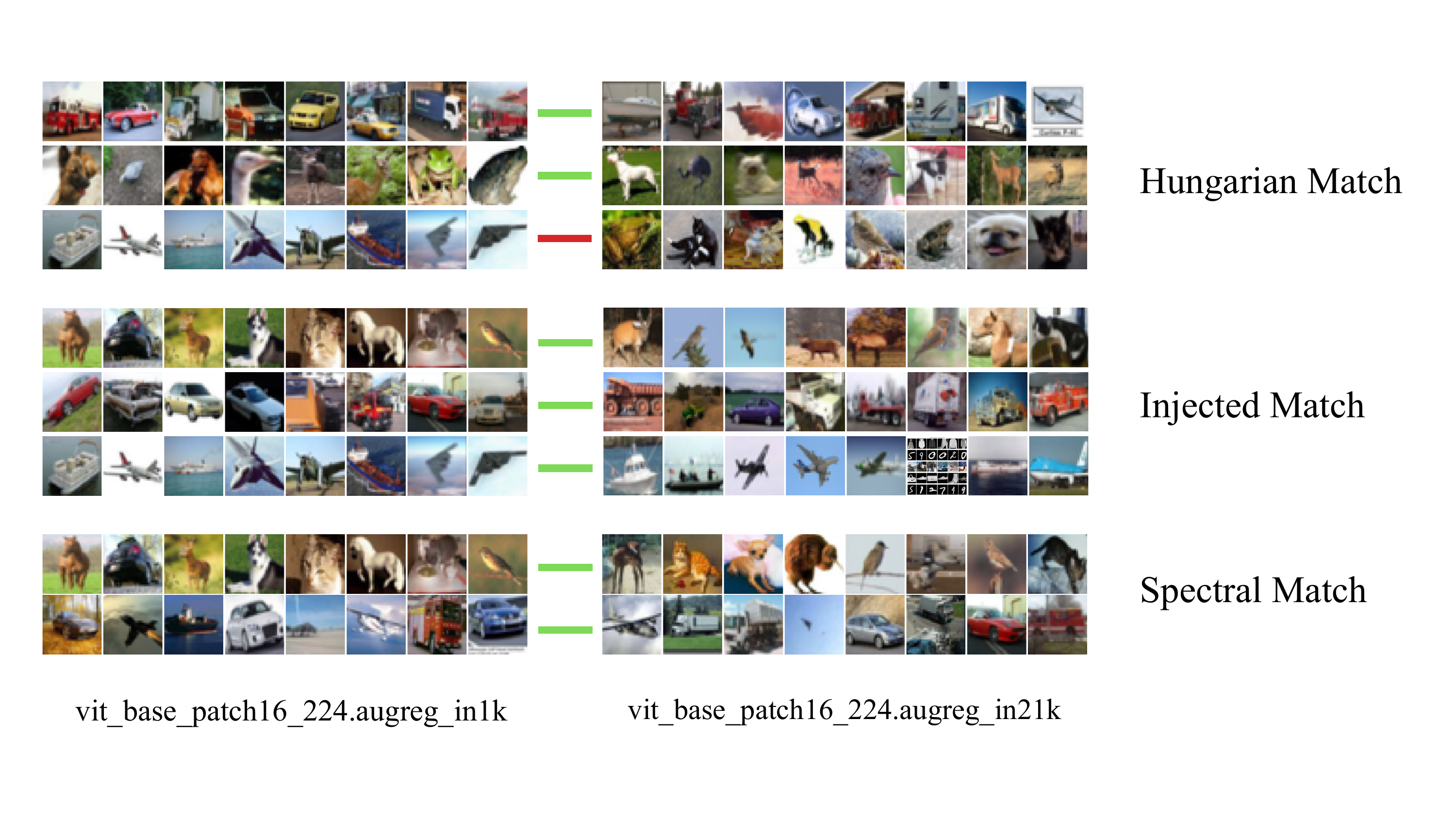}
    \caption{Prototype correspondences between \texttt{vit\_base\_patch16\_224.augreg\_in1k} 
    (left) and \texttt{vit\_base\_patch16\_224.augreg\_in21k} (right) on CIFAR-10. 
    Each row represents a matched pair of clusters, with green connectors indicating 
    semantically coherent correspondences and red connectors indicating mismatches. 
    The three panels show results for Hungarian matching (top), injected matching 
    (middle), and spectral matching (bottom).}
    \label{fig:matching_cifar10}
\end{figure}

Figure~\ref{fig:matching_cifar10} shows results on CIFAR-10, a dataset consisting of images spanning animals, vehicles, and aircraft. The three matching methods exhibit markedly different behaviors. Hungarian matching produces four green-connected pairs and one red pair. The green pairs capture semantically coherent correspondences, correctly aligning clusters of animals and vehicles across the two models. The single red pair associates a cluster dominated by aircraft and watercraft in $\mathcal{A}$ with a cluster of small animals in $\mathcal{B}$. This mismatch is not incidental but rather reflects a structural limitation of the one-to-one assignment constraint: the transport-related concepts encoded by $\mathcal{A}$ in a single prototype are distributed across multiple clusters in $\mathcal{B}$, and the Hungarian algorithm, being unable to capture such many-to-one correspondences, is forced into a semantically incoherent pairing. Injected matching yields five green pairs, all consistent by construction since the partition is entirely determined by $\mathcal{A}$. Spectral matching recovers three green pairs with broader, semantically compact groups. The reduced number of correspondences suggests that the method has identified many-to-many structures, merging prototypes that the two models distribute differently across their respective partitions.

\begin{figure}[h]
    \centering
    \includegraphics[width=\textwidth]{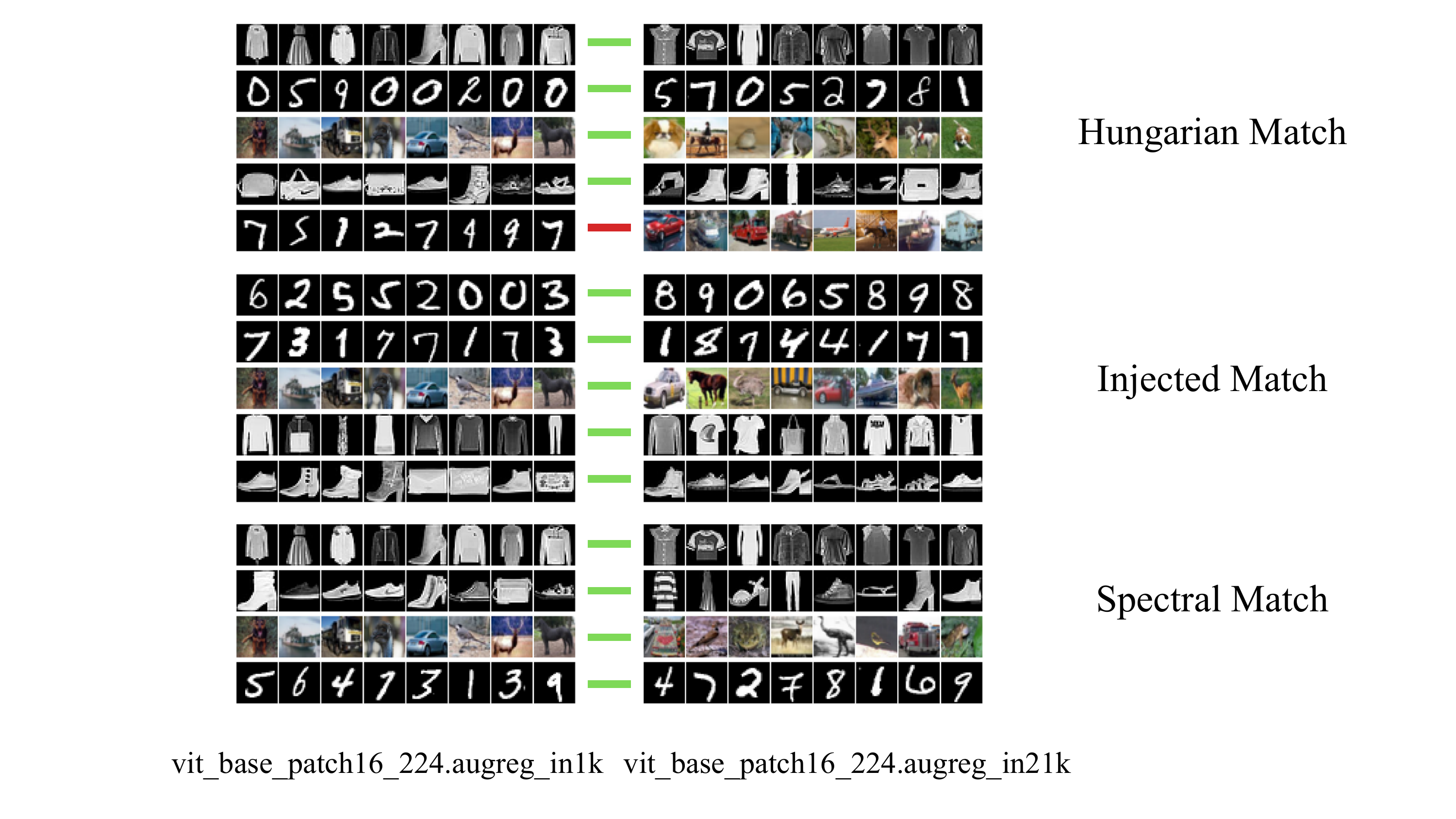}
    \caption{Prototype correspondences between \texttt{vit\_base\_patch16\_224.augreg\_in1k} 
    (left) and \texttt{vit\_base\_patch16\_224.augreg\_in21k} (right) on a multi-dataset 
    benchmark combining CIFAR-10, MNIST, and Fashion-MNIST. Each row represents a matched 
    pair of clusters, with green connectors indicating semantically coherent correspondences 
    and red connectors indicating mismatches. At this scale, individual datasets act as 
    macro-concepts, allowing the matching methods to be evaluated on their ability to recover 
    coarse semantic partitions. The three panels show results for Hungarian matching (top), 
    injected matching (middle), and spectral matching (bottom)}
    \label{fig:matching_multi}
\end{figure}

Figure~\ref{fig:matching_multi} shows results on a multi-dataset benchmark obtained by combining CIFAR-10, MNIST, and Fashion-MNIST, thus covering a diverse range of visual domains including natural images, handwritten digits, and clothing items. When operating at this scale, individual datasets emerge as macro-concepts, and the matching methods can be evaluated on their ability to recover these coarse semantic partitions. Hungarian matching produces four green pairs and one red pair. The green pairs reveal that both models consistently separate the three domains, correctly aligning clusters of clothing items (Fashion-MNIST), handwritten digits (MNIST), and natural images (CIFAR-10) across the two encoders. The single red pair, however, exposes the structural limitation of the one-to-one assignment constraint: the digit macro-concept is distributed across a different number of prototypes in $\mathcal{A}$ and $\mathcal{B}$, and the Hungarian algorithm, being unable to capture such many-to-many correspondences, is forced to match a digit cluster in $\mathcal{A}$ with a natural image cluster in $\mathcal{B}$. Injected matching yields five green pairs, all consistent by construction, but the clusters attributed to $\mathcal{B}$ are more heterogeneous, with digits and natural images occasionally co-occurring within the same group. Spectral matching recovers four green pairs with semantically compact groups, cleanly separating clothing, digits, and natural image categories, confirming that both models agree on the macro-concept level partition induced by the dataset composition, while naturally handling the many-to-many structure that Hungarian matching fails to capture.
\section{Semantic Alignment Methodologies}
\label{app:alignment}

\subsection{The Semantic Alignment problem}

The \textbf{semantic alignment} problem arises when two independently trained models produce representations of the same domain in mutually incompatible latent spaces. Given a source model $A$ and a target model $B$, the goal is to find a map $\mathcal{T}: \mathcal{Z}_A \to \mathcal{Z}_B$ such that the transferred embeddings are semantically comparable to those of $B$, without requiring joint retraining or access to raw data at inference time.
 
This \textit{semantic mismatch} is not merely a technical inconvenience: embeddings from different architectures or training regimes may differ not only in dimensionality but also in the orientation, scale, and curvature of their latent geometry. A classifier or retrieval system trained on $B$'s representations will therefore fail systematically when fed embeddings from $A$, even if the two models encode the same semantic content. Alignment methods bridge this gap by learning a data-driven transformation on a shared set of paired training embeddings $\{(a_i, b_i)\}_{i=1}^n$, where $a_i \in \mathcal{Z}_A$ and $b_i \in \mathcal{Z}_B$ correspond to the same input sample.

A rich body of work has explored this problem from multiple perspectives, including anchor-based relative representations~\citep{moschella2023relative,fiorellino2024dynamic,huttebraucker2024relative}, supervised linear mappings~\citep{merullo2022linearly,moayeri2023text,maiorca2023latent,lahner2024direct,pandolfo2025latent,di2025federated,grimaldi2025learning}, spectral and geometric approaches~\citep{fumero2024latent,grimaldi2025learning}, and optimal transport or contrastive learning methods~\citep{zhang2017earth,alvarez2018gromov,grave2019unsupervised,jha2025harnessing}.

\subsection{Semantic Alignment Pipeline}

\begin{figure}[t]
    \centering
    \includegraphics[width=\textwidth, trim={0 12cm 0 0 0}]{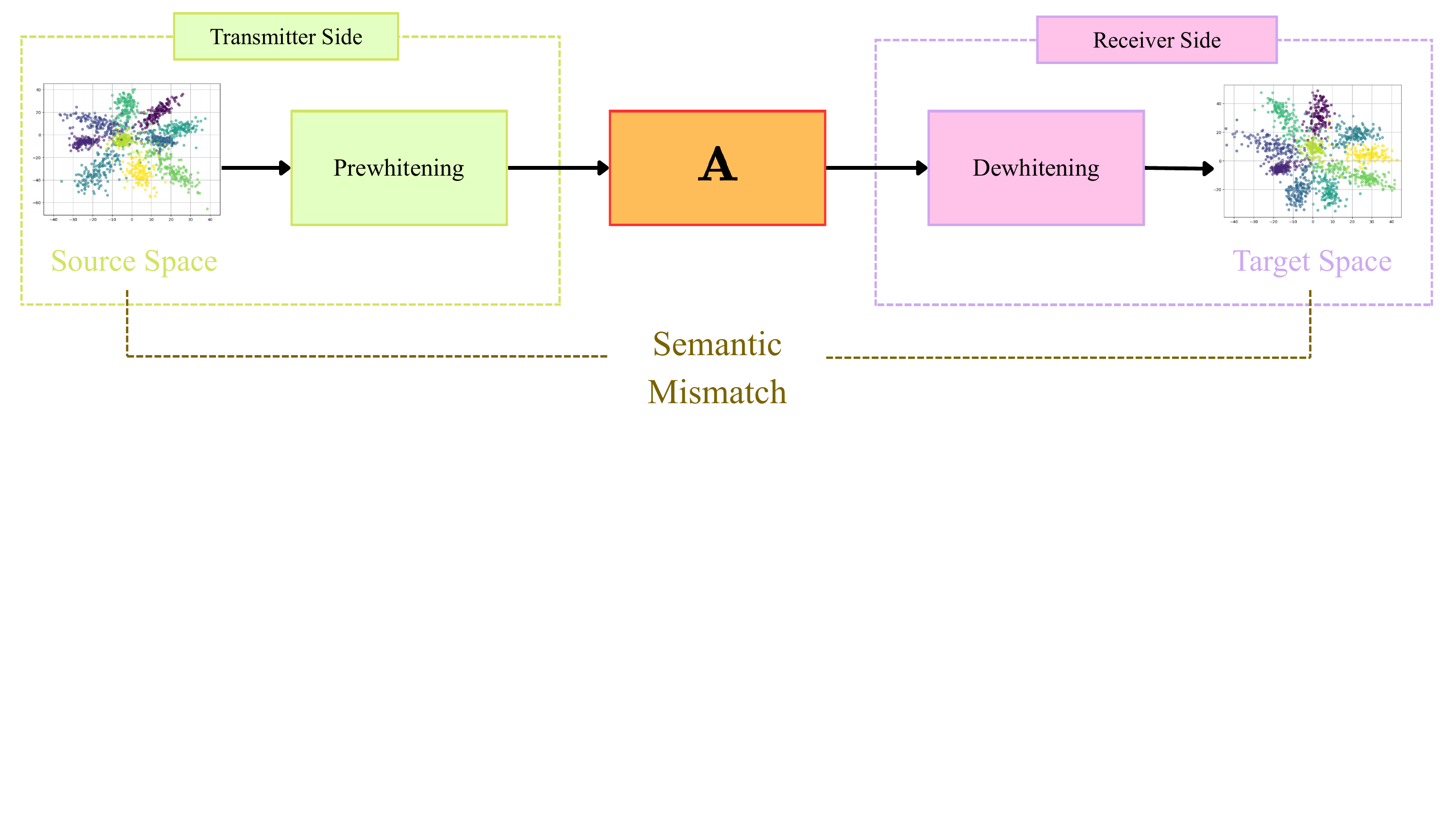}
    \caption{The three-stage alignment pipeline shared by all methods considered
    in this work. A test embedding $\mathbf{a} \in \mathcal{Z}_A$ is first
    \emph{prewhitened} into a canonical coordinate system via $\mathbf{W} = \mathbf{L}^{-1}$
    (Cholesky-based whitening), then mapped by the method-specific alignment
    operator $\mathbf{A}$, and finally \emph{dewhitened} back into
    $\mathcal{Z}_B$ via $\mathbf{W}^{-1} = \mathbf{L}$, yielding the transmitted embedding
    $\hat{\mathbf{b}} \in \mathcal{Z}_B$. The normalisation steps are shared across all
    methods; only $\mathbf{A}$ varies.}
    \label{fig:pipeline}
\end{figure}

Most of the methods considered in this work share the same three-stage pipeline, which decomposes the alignment into a \emph{normalisation} step, a \emph{linear transformation} step, and an \emph{inverse normalisation} step. Given a test embedding $\mathbf{a} \in \mathcal{Z}_A$, the transmitted embedding $\hat{\mathbf{b}} \in \mathcal{Z}_B$ is obtained as illustrated in Figure~\ref{fig:pipeline}.

The matrix $\mathbf{A}$ is the alignment operator, whose specific form depends on the method (see Appendix \ref{apps:alignment-methodologies}). The prewhitening and dewhitening steps are shared across methods and are described below.

\textbf{Prewhitening.} Whitening is a linear normalisation that maps a set of embeddings to a canonical coordinate system with zero mean and identity covariance. It serves two purposes: (i) it removes the influence of the individual scale and correlation structure of each space, making the alignment problem more symmetric; (ii) it stabilises the estimation of the alignment operator by conditioning the data.

Concretely, let $\mathbf{X} \in \mathbb{R}^{n \times d}$ be the matrix of training embeddings with mean $\boldsymbol{\mu}$ and empirical covariance
\[
    \mathbf{C} = \frac{1}{n-1}\, \mathbf{X}_c^\top \mathbf{X}_c + \varepsilon \mathbf{I},
    \qquad \mathbf{X}_c = \mathbf{X} - \mathbf{1}\boldsymbol{\mu}^\top,
\]
where $\varepsilon = 10^{-6}$ ensures positive definiteness. The covariance is factorised via its Cholesky decomposition $\mathbf{C} = \mathbf{L}\mathbf{L}^\top$, where $\mathbf{L} \in \mathbb{R}^{d \times d}$ is lower triangular. The whitening operator is $\mathbf{W} = \mathbf{L}^{-1}$, and the whitened embeddings are
\[
    \widetilde{\mathbf{X}} = \mathbf{X}_c\, \mathbf{W}^\top = \mathbf{X}_c\, \mathbf{L}^{-\top}.
\]
The result $\widetilde{\mathbf{X}}$ has covariance approximately equal to the identity matrix, i.e.\ its dimensions are decorrelated and each has unit variance. Both $\mathbf{L}$ and $\boldsymbol{\mu}$ are stored at training time for use at inference. The prewhitening is applied independently to both spaces $\mathcal{Z}_A$ and $\mathcal{Z}_B$, yielding $\widetilde{\mathbf{A}}$ and $\widetilde{\mathbf{B}}$ respectively.

\textbf{Alignment operator.} After prewhitening, the alignment operator $\mathbf{A} \in \mathbb{R}^{d_A \times d_B}$ (or more generally a map $\mathbb{R}^{d_A} \to \mathbb{R}^{d_B}$) is estimated from the paired whitened embeddings $(\widetilde{\mathbf{A}}, \widetilde{\mathbf{B}})$.
The specific form of $\mathbf{A}$ --- whether a prototype frame, a truncated linear map, or a canonical projection --- defines the alignment method and determines the transmitted whitened embedding $\widetilde{\mathbf{b}} = \mathbf{A}\,\widetilde{\mathbf{a}}$.

\textbf{Dewhitening.} Dewhitening is the inverse of the prewhitening applied to $\mathcal{Z}_B$. It maps the transmitted whitened embedding back to the original coordinate system of $\mathcal{Z}_B$:
\[
    \hat{\mathbf{b}} = \widetilde{\mathbf{b}}\, \mathbf{L}_B^\top + \boldsymbol{\mu}_B,
\]
where $\mathbf{L}_B$ is the Cholesky factor of $\mathcal{Z}_B$'s empirical covariance and $\boldsymbol{\mu}_B$ is its mean. Geometrically, $\mathbf{L}_B$ acts as a square root of the covariance of $\mathcal{Z}_B$: the map $\mathbf{z} \mapsto \mathbf{z}\mathbf{L}_B^{-\top}$ whitens (decorrelates and normalises), while its inverse $\mathbf{z} \mapsto \mathbf{z}\mathbf{L}_B^\top$ re-introduces the original covariance structure. The dewhitening step ensures that $\hat{\mathbf{b}}$ lies in the same geometric space as the embeddings of $\mathcal{Z}_B$, making it directly compatible with any downstream model trained on $\mathcal{Z}_B$.

\subsection{Alignment Methodologies}
\label{apps:alignment-methodologies}

We evaluate three alignment methods of increasing structural complexity. All
operate on the prewhitened spaces $\widetilde{\mathbf{A}}$ and $\widetilde{\mathbf{B}}$, and all
are parametrised by a rank/dimensionality hyperparameter $k$ that controls the
effective complexity of the transmitted representation. The methods differ in
the inductive bias they impose on the alignment operator $\mathbf{A}$.

\subsubsection{ \textsc{Proto} --- Prototype-based Parseval Frame}

This method exploits the geometric structure of both latent spaces through a framework of \emph{prototypes} and \emph{Parseval Frame Equalizers} \cite{fiorellino2025frame}. Prototypical anchors are computed from $\mathcal{Z}_A$'s whitened embeddings via Algorithm~\ref{alg:proto_alg}, producing $k$ centroids and a shared index set $\mathcal{A}$. Let $\widetilde{\mathbf{X}}_\mathcal{A} \in \mathbb{R}^{|\mathcal{A}| \times d_A}$ and $\widetilde{\mathbf{Y}}_\mathcal{A} \in \mathbb{R}^{|\mathcal{A}| \times d_B}$ denote the whitened anchor embeddings of $\mathcal{Z}_A$ and $\mathcal{Z}_B$ respectively, indexed by $\mathcal{A}$. The private PFE operators are
\[
    \mathbf{F}_T \;=\; \widetilde{\mathbf{X}}_\mathcal{A}
        \!\left(\widetilde{\mathbf{X}}_\mathcal{A}^\top
        \widetilde{\mathbf{X}}_\mathcal{A}\right)^{-1/2},
    \qquad
    \mathbf{F}_R \;=\; \widetilde{\mathbf{Y}}_\mathcal{A}
        \!\left(\widetilde{\mathbf{Y}}_\mathcal{A}^\top
        \widetilde{\mathbf{Y}}_\mathcal{A}\right)^{-1/2},
\]
where $\mathbf{F}_T$ is the \emph{analysis operator} of $\mathcal{Z}_A$ and $\mathbf{F}_R^\top$ is the \emph{synthesis operator} of $\mathcal{Z}_B$. The normalisation by $(\widetilde{\mathbf{X}}_\mathcal{A}^\top \widetilde{\mathbf{X}}_\mathcal{A})^{-1/2}$ ensures that both operators satisfy the Parseval condition $\mathbf{F}_T^\top \mathbf{F}_T = \mathbf{F}_R^\top \mathbf{F}_R = \mathbf{I}$~\cite{fiorellino2025frame}, which guarantees norm-preserving projections. The alignment operator is then obtained by composing the two:
\[
    \mathbf{A} \;=\; \mathbf{F}_R^\top \mathbf{F}_T \;\in\; \mathbb{R}^{d_B \times d_A},
\]
so that the transmitted whitened embedding is $\widetilde{\mathbf{b}} = \mathbf{A}\,\widetilde{\mathbf{a}} \in \mathbb{R}^{d_B}$, which is subsequently dewhitened to obtain $\hat{\mathbf{b}}$. The anchor embeddings of $\mathcal{Z}_B$ are computed via the \emph{injected matching} scheme (Appendix~\ref{app:compare_concepts}): rather than clustering $\{\widetilde{\mathbf{b}}_i\}$ independently, the same index set $\mathcal{A}$ is reused, forcing a shared semantic partition across both spaces.

\subsubsection{\textsc{Linear} --- Rank-$k$ Truncated Linear Map}

 This method learns an optimal linear map between the two whitened spaces and uses a low-rank approximation to control the effective dimensionality of the transmission. Unlike the prototype approach, no structural prior is imposed on the map; the solution is purely data-driven.
 
\textbf{Least-squares map.} Let $\mathbf{Z}_A \in \mathbb{R}^{d_A \times n}$ and $\mathbf{Z}_B \in \mathbb{R}^{d_B \times n}$ be the matrices collecting the $n$ whitened training embeddings of $\mathcal{Z}_A$ and $\mathcal{Z}_B$ respectively. The alignment operator $\mathbf{A} \in \mathbb{R}^{d_B \times d_A}$ is obtained by solving
\[
    \mathbf{A} \;=\; \arg\min_{\mathbf{A}}\,\bigl\|\mathbf{Z}_B - \mathbf{A}\,\mathbf{Z}_A\bigr\|_F^2,
\]
whose closed-form solution is $\mathbf{A} = \mathbf{Z}_B\,\mathbf{Z}_A^{\dagger}$.
 
\textbf{Rank-$k$ truncation via SVD.} The operator $\mathbf{A}$ is decomposed as $\mathbf{A} = \mathbf{U}\,\boldsymbol{\Sigma}\,\mathbf{V}^\top$. For a given rank $k$, the truncated operator
\[
    \mathbf{A}_k \;=\; \mathbf{U}_{:k}\,\boldsymbol{\Sigma}_k\,\mathbf{V}_{:k}^\top
\]
retains only the $k$ most informative directions. The transmitted whitened embedding is
\[
    \widetilde{\mathbf{b}} \;=\; \mathbf{A}_k\,\widetilde{\mathbf{a}} \;\in\; \mathbb{R}^{d_B}.
\]
The SVD is computed \emph{once} on the full operator $\mathbf{A}$ and the truncation is applied separately for each value of $k$, making \textsc{Linear} the most computationally efficient method to sweep over $k$.

\subsubsection{\textsc{CCA} --- Canonical Correlation Analysis}

Canonical Correlation Analysis finds pairs of linear projections --- one for each space --- that maximise the correlation between the projected embeddings. The resulting $k$-dimensional canonical space serves as a shared intermediate representation through which the transmission is performed. Differently from \textsc{Proto} and \textsc{Linear}, CCA operates directly on the \emph{raw} embeddings and \emph{bypasses both the prewhitening and dewhitening steps}: whitening is implicit in the construction of the canonical directions, and re-centering plays the role of dewhitening at inference time.
 
\textbf{Covariance estimation.} Let $\mathbf{X}_c = \mathbf{A}_{\text{train}} - \boldsymbol{\mu}_A$ and $\mathbf{Y}_c = \mathbf{B}_{\text{train}} - \boldsymbol{\mu}_B$ be the centred training matrices. The empirical covariance and cross-covariance matrices are
\[
    \mathbf{S}_{AA} = \frac{\mathbf{X}_c^\top \mathbf{X}_c}{n-1} + \varepsilon \mathbf{I}, \qquad
    \mathbf{S}_{BB} = \frac{\mathbf{Y}_c^\top \mathbf{Y}_c}{n-1} + \varepsilon \mathbf{I}, \qquad
    \mathbf{S}_{AB} = \frac{\mathbf{X}_c^\top \mathbf{Y}_c}{n-1}.
\]
 
\textbf{Canonical directions.} The cross-whitened matrix
\[
    \mathbf{T} \;=\; \mathbf{S}_{AA}^{-1/2}\,\mathbf{S}_{AB}\,\mathbf{S}_{BB}^{-1/2}
\]
is decomposed via SVD as $\mathbf{T} = \mathbf{U}\,\boldsymbol{\Sigma}\,\mathbf{V}^\top$. The canonical projection
matrices are
\[
    \mathbf{W}_A \;=\; \mathbf{S}_{AA}^{-1/2}\,\mathbf{U}_{:k} \;\in\; \mathbb{R}^{d_A \times k},
    \qquad
    \mathbf{W}_B \;=\; \mathbf{S}_{BB}^{-1/2}\,\mathbf{V}_{:k} \;\in\; \mathbb{R}^{d_B \times k},
\]
where $\mathbf{S}_{AA}^{-1/2}$ and $\mathbf{S}_{BB}^{-1/2}$ are obtained via the spectral
decomposition of the respective covariance matrices.
 
\textbf{Transmission.}
A test embedding $\mathbf{a}$ is taken in its \emph{raw} form --- no prewhitening is applied --- projected into the canonical space, and then lifted back to $\mathcal{Z}_B$'s original space via the pseudo-inverse of $\mathbf{W}_B$:
\[
    \mathbf{z} \;=\; (\mathbf{a} - \boldsymbol{\mu}_A)\,\mathbf{W}_A, \qquad
    \hat{\mathbf{b}} \;=\; \mathbf{z}\,\mathbf{W}_B^{\dagger} + \boldsymbol{\mu}_B.
\]
The re-centering by $\boldsymbol{\mu}_B$ plays the role of the dewhitening step, so no separate Cholesky-based dewhitening is required. CCA therefore departs from the three-stage pipeline described above: it implicitly normalises both spaces through $\mathbf{T}$ and handles centering at both ends, making prewhitening and dewhitening redundant.

\subsection{Alignment Evaluation}
 
We evaluate the quality of the alignment along two complementary axes: reconstruction fidelity and downstream task performance.
 
\textbf{Evaluation protocol.} Once the alignment operator $\mathbf{A}$ has been estimated on the training set, model $\mathcal{Z}_A$ transmits its held-out test embeddings to $\mathcal{Z}_B$ through the three-stage pipeline. Concretely, for each test embedding $\mathbf{a} \in \mathcal{Z}_A$, the transmitted embedding $\hat{\mathbf{b}}$ is obtained as
\[
    \mathbf{a}
    \;\xrightarrow{\text{prewhitening}}\;
    \widetilde{\mathbf{a}}
    \;\xrightarrow{\;\mathbf{A}\;}\;
    \widetilde{\mathbf{b}}
    \;\xrightarrow{\text{dewhitening}}\;
    \hat{\mathbf{b}},
\]
and is then compared against the corresponding ground-truth embedding $\mathbf{b} \in \mathcal{Z}_B$, i.e.\ the representation that model $\mathcal{Z}_B$ would have produced for the same input.
 
\textbf{Reconstruction fidelity.} The mean squared error between the transmitted and the ground-truth embeddings is computed over the test set of $n_{\text{test}}$ samples:
\[
    \mathrm{MSE} \;=\; \frac{1}{n_{\text{test}}} \sum_{i=1}^{n_{\text{test}}}
    \bigl\|\hat{\mathbf{b}}_i - \mathbf{b}_i\bigr\|_2^2.
\]
This metric directly quantifies the geometric distortion introduced by the alignment, independently of any downstream task.
 
\textbf{Downstream performance.} To assess whether the aligned embeddings are semantically usable, we evaluate classification accuracy via linear probing. A linear classifier trained on $\mathcal{Z}_B$'s training embeddings is applied to the transmitted test embeddings $\hat{\mathbf{b}}$, and the resulting accuracy is compared to the upper bound obtained by probing $\mathcal{Z}_B$'s own test embeddings $\mathbf{b}$. The gap between the two measures the cost of alignment in terms of task-relevant information.

\section{Statistical Analysis}
\label{app:stat_regression}

For each of the five treatments, we fit a pooled OLS regression per target metric with HC3 heteroskedasticity-consistent standard errors. Most of the dependent variables are described in Appendix~\ref{app:target_variables}. The analysis spans four evaluation datasets: CIFAR-10, MNIST, Fashion-MNIST, and Oxford Flowers. The treatment indicator $\mathbf{1}[\text{treated}]_i$ is binary (0 = control, 1 = treatment) and is built from matched pairs: each pair contributes one control row and one treatment row sharing the same architecture family. Formally, for each metric $y$ we estimate
\[
y_{i} = \alpha + \beta \cdot \mathbf{1}[\text{treated}]_i + \gamma^\top \mathbf{a}_i +
\delta^\top \mathbf{d}_i + \varepsilon_i,
\]
where $\mathbf{a}_i$ and $\mathbf{d}_i$ are vectors of architecture-family and evaluation-dataset fixed effects respectively, included as nuisance controls. The coefficient of interest is $\beta$, which answers: \textit{does the treatment affect this metric, regardless of which dataset or architecture family?}
 
Each treatment is constructed as a strict \textit{ceteris paribus} contrast, isolating a single pretraining factor by holding all others fixed. For the first three treatments, Dataset Complexity, Specialization, and Transfer Learning, we rely on ImageNet variants (IN-1K and IN-21K) as the pretraining datasets. This choice is deliberate: ImageNet variants have a well-defined ordering in terms of scale and informativeness, allowing us to unambiguously establish which dataset constitutes the more complex pretraining condition without relying on heuristic arguments.
\begin{itemize}
    \item \textbf{Dataset Complexity} varies only the pretraining dataset, comparing a smaller ImageNet variant (IN-1K) against a larger one (IN-21K). Architecture, augmentation strategy, and training procedure are identical across the pair.
 
    \item \textbf{Specialization} varies only whether fine-tuning has occurred. The control is a model pretrained on IN-21K without any subsequent fine-tuning; the treatment is the same model checkpoint further fine-tuned to IN-1K. Architecture, augmentation, and base
    pretraining are therefore shared.
 
    \item \textbf{Transfer Learning} varies the pretraining source while holding the final training target fixed. Both the control and treatment are evaluated after training on IN-1K, but the control was trained on IN-1K directly, whereas the treatment was first pretrained on IN-21K and then fine-tuned to IN-1K. Architecture and augmentation are held constant.
 
    \item \textbf{Augmentation} varies only whether data augmentation was applied during pretraining. Both models share the same architecture, the same pretraining dataset (IN-21K), and the same training procedure, differing solely in the use of augmentation regularisation.
 
    \item \textbf{Model Scale} varies only the size of the model within a fixed architectural family. The control is a smaller variant (e.g., ViT-Small) and the treatment is a larger variant (e.g., ViT-Base), trained on the same dataset with the same augmentation and setup.
\end{itemize}
 
This design ensures that any systematic difference in the geometric metrics can be attributed to the manipulated factor rather than to confounds. Table~\ref{tab:conditions} details the design of each treatment.

\begin{table}[t]
\centering
\small
\caption{Design of the five pretraining conditions. For each condition, all factors not listed are held fixed. Examples are drawn from actual models in SEMASIA.}
\label{tab:conditions}
\vspace{6pt}
\setlength{\tabcolsep}{5pt}
\begin{tabular}{p{3.2cm} p{3.8cm} p{3.8cm} r}
\textbf{Analysis Type} & \textbf{Control} & \textbf{Treatment} & \textbf{N. Obs.} \\
\midrule
Dataset Complexity
    & Smaller ImageNet variant \newline {\scriptsize\ttfamily vit\_base\_patch16\_224.augreg\_in1k}
    & Larger ImageNet variant \newline {\scriptsize\ttfamily vit\_base\_patch16\_224.augreg\_in21k}
    & 328 \\
\addlinespace
Specialization
    & Same model, no fine-tuning \newline {\scriptsize\ttfamily vit\_base\_patch16\_224.augreg\_in21k}
    & Same model fine-tuned to ImageNet-1K \newline {\scriptsize\ttfamily vit\_base\_patch16\_224.augreg\_in21k\_ft\_in1k}
    & 484 \\
\addlinespace
Transfer Learning
    & Pretrained directly on ImageNet-1K \newline {\scriptsize\ttfamily vit\_base\_patch16\_224.augreg\_in1k}
    & Large pretraining $+$ fine-tuned to ImageNet-1K \newline {\scriptsize\ttfamily vit\_base\_patch16\_224.augreg\_in21k\_ft\_in1k}
    & 452 \\
\addlinespace
Augmentation
    & Same model, same pretraining dataset, no augmentation \newline {\scriptsize\ttfamily vit\_base\_patch16\_224.orig\_in21k}
    & Same model, same pretraining dataset, with augmentation \newline {\scriptsize\ttfamily vit\_base\_patch16\_224.augreg\_in21k}
    & 224 \\
\addlinespace
Model Scale
    & Same architectural family, smaller model variant \newline {\scriptsize\ttfamily vit\_small\_patch16\_224.augreg\_in1k}
    & Same architectural family, larger model variant \newline {\scriptsize\ttfamily vit\_base\_patch16\_224.augreg\_in1k}
    & 7{,}260 \\
\bottomrule
\end{tabular}
\end{table}

\subsection{Target Variables}
\label{app:target_variables}

\textbf{Total Spread}. Given an embedding $Z \in \mathbb{R}^{n \times d}$, we measure its global scale using its total variance. Mathematically, this is defined as the trace of the covariance matrix,
\[
TS(Z)=\mathrm{Tr}(\mathrm{Cov}(Z)) = \sum_{j=1}^d \mathrm{Var}(Z_j).
\]

This quantity captures the overall spread of the representation around its mean, aggregating variance across all dimensions. 

\textbf{Mean Distance to Centroid}. Given an embedding $Z \in \mathbb{R}^{n \times d}$, we consider the mean Euclidean distance of points from the centroid as an additional measure of scale,
\[
MDC(Z)=\frac{1}{n} \sum_{i=1}^n \|z_i - \mu\|,
\]
where $\mu$ is the mean of the embedding.

Compared to total spread, which averages squared distances, this quantity is less sensitive to outliers, as it grows linearly with the distance from the centroid.

\textbf{Standard Deviation of Distance to Centroid}. Let $Z \in \mathbb{R}^{n \times d}$ be an embedding. To assess how uniformly its points are distributed, we consider the standard deviation of their Euclidean distances to the centroid:
\[
SDDC(Z)= \sqrt{
\frac{1}{n} \sum_{i=1}^n \|z_i - \mu\|^2
-
\left(
\frac{1}{n} \sum_{i=1}^n \|z_i - \mu\|
\right)^2
} \]

While mean distance captures the typical scale of the embedding, this quantity measures how consistent that scale is across points. Low values indicate that points lie at similar distances from the centroid, suggesting a uniform spread, whereas high values reflect heterogeneity, such as the presence of clusters or outliers.

\textbf{Density Estimate}. Given an embedding $Z \in \mathbb{R}^{n \times d}$, we define a simple proxy for its global density as the ratio between the number of points and the total variance,
\[
\rho(Z) = \frac{n}{\mathrm{Tr}(\mathrm{Cov}(Z))}.
\]

This quantity measures how many points are packed within the overall spread of the representation. Higher values indicate that points are more concentrated in a smaller region of the latent space, while lower values correspond to more diffuse embeddings.

\textbf{Number of Components for 90\% Variance}. Given an embedding $Z \in \mathbb{R}^{n \times d}$, we estimate its dimensionality using principal component analysis. Let $\{\lambda_i\}$ denote the eigenvalues of the covariance matrix in decreasing order. We define
\[
k_{0.9}(Z) = \min \left\{ k : \frac{\sum_{i=1}^k \lambda_i}{\sum_{i=1}^d \lambda_i} \ge 0.9 \right\},
\]
the number of principal components required to explain 90\% of the total variance.

Lower values indicate that most of the variance is captured by a small number of directions, while higher values reflect a more distributed representation.

\textbf{Explained Variance Ratios}. Given an embedding $Z \in \mathbb{R}^{n \times d}$, we analyze how variance is concentrated in the leading principal components. Let $\{\lambda_i\}$ denote the eigenvalues of the covariance matrix in decreasing order, and define the explained variance ratios as $r_i = \lambda_i / \sum_j \lambda_j$.

We report the proportion of variance explained by the top principal components,
\[
EVR_1(Z)=r_1 \quad \text{and} \quad EVR_3(Z)=\sum_{i=1}^3 r_i,
\]
corresponding to the top-1 and top-3 components, respectively.

These quantities capture how strongly the representation is dominated by a small number of directions. High values indicate that most of the variance lies in a few principal components, while lower values reflect a more distributed structure.

\textbf{Isotropy}. Given an embedding $Z \in \mathbb{R}^{n \times d}$, we quantify isotropy by comparing the minimum and maximum variance across dimensions,
\[
Is(Z)=\frac{\min_j \mathrm{Var}(Z_j)}{\max_j \mathrm{Var}(Z_j)}.
\]

This provides a simple proxy for how uniformly variance is distributed across coordinates. Values close to 1 indicate isotropic representations, while values near 0 reflect strong anisotropy, with variance concentrated in a few directions.

\textbf{Spectral Entropy}. To quantify how information is distributed across the latent space, we consider the spectral entropy of the embedding matrix \cite{shannon1948}. Given an embedding $Z \in \mathbb{R}^{n \times d}$, we first center it and compute its singular values $\{\sigma_i\}$. These are then normalized to form a probability distribution
\[
p_i = \frac{\sigma_i}{\sum_j \sigma_j}.
\]
The spectral entropy is defined as
\[
H(Z) = -\sum_i p_i \log p_i.
\]

Intuitively, this quantity measures how evenly the variance of the representation is spread across different directions. When most of the mass is concentrated in a few singular values, the entropy is low, indicating that the embedding effectively lies in a low-dimensional subspace. Conversely, when the singular values are more uniform, the entropy is higher, reflecting a more isotropic representation.


\textbf{Effective Rank}. To obtain a real-valued measure of dimensionality, we use the effective rank of the embedding as the exponential of its spectral entropy \cite{roy2007effective}. Given a centered embedding $Z \in \mathbb{R}^{n \times d}$, 
the effective rank is then defined as
\[
r_{\mathrm{eff}}(Z) = \exp(H(Z)).
\]

This quantity can be interpreted as the number of dimensions that are effectively used by the representation. In particular, if the singular values are uniformly distributed over $k$ directions, then $H(Z) = \log k$ and $r_{\mathrm{eff}}(Z) = k$. More generally, the effective rank provides a smooth proxy for dimensionality, taking non-integer values when variance is unevenly distributed across directions.

Compared to the algebraic rank, which is sensitive to noise and thresholding, the effective rank captures the intrinsic structure of the embedding by accounting for how variance is distributed across its principal components.

\textbf{Linear Probing Metrics}. To quantify how much class information is encoded in the learned representations, we perform a linear probing task. Given an embedding $Z \in \mathbb{R}^{n \times d}$ and class labels $y_i \in \{1,\dots,C\}$, a linear classifier
\[
f(z) = Wz + b
\]
is trained to predict the dataset class labels from the embedding vectors. High predictive performance indicates that classes are linearly separable in the latent space.

We report four standard classification metrics: accuracy, precision, recall, and F1-score \cite{manning2008ir}. Accuracy measures the overall fraction of correctly classified samples. Precision evaluates how often predicted labels are correct, while recall measures how many true instances of each class are successfully recovered. The F1-score is the harmonic mean of precision and recall, balancing both aspects of performance.

All metrics are computed using macro-averaging across classes, so that each dataset class contributes equally regardless of class frequency.
\begin{figure}[t]
    \centering
    \includegraphics[width=\textwidth]{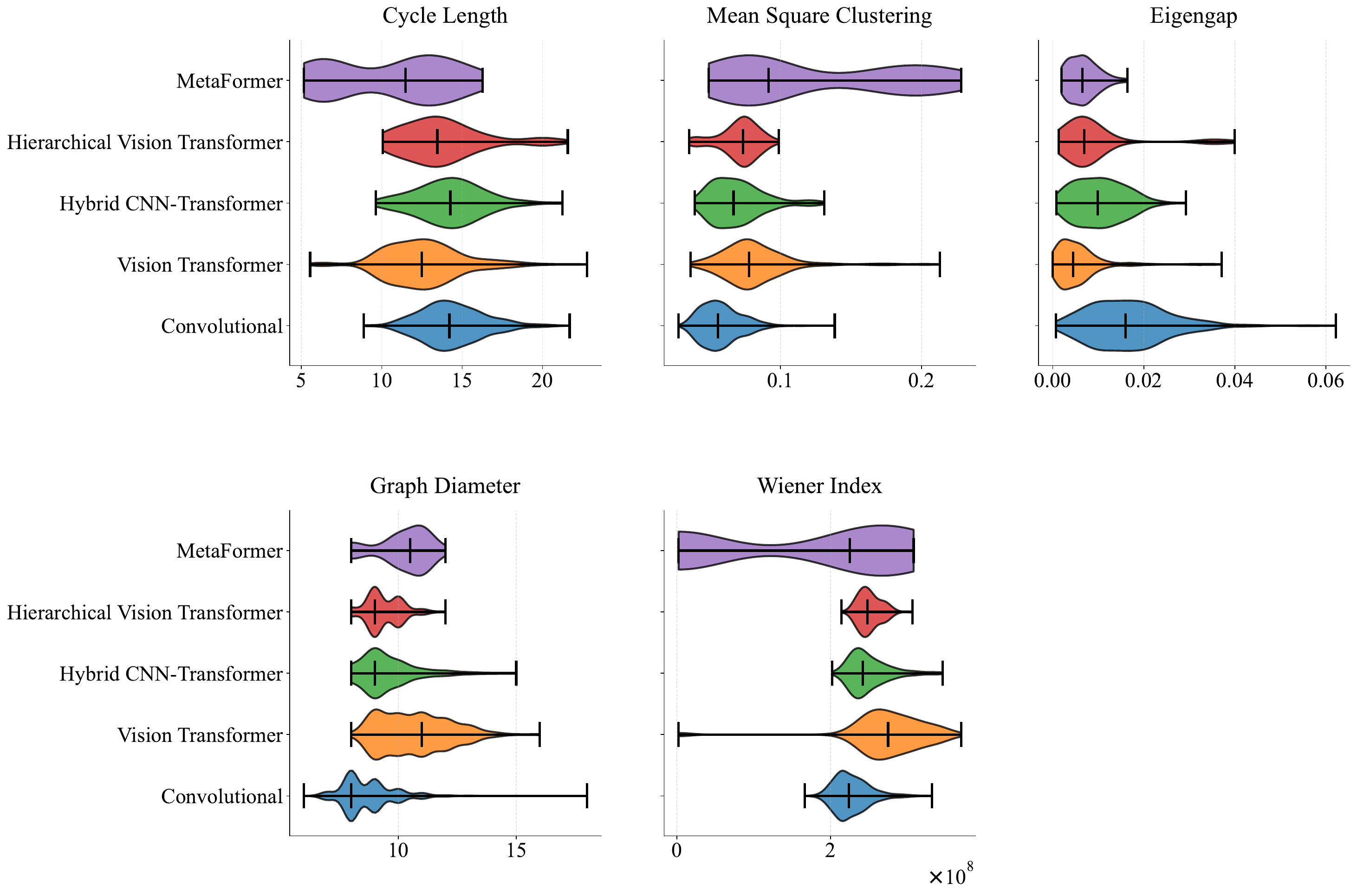}
    \caption{Violin plots of five latent graph signatures across model macro-families on CIFAR-10. For each model, embeddings are represented as a point cloud and converted into a $k$-nearest-neighbor graph ($k=10$), from which the reported descriptors are computed. 
    }
    \label{fig:graph_signatures}
\end{figure}

\section{Graph-Based Regression Analysis}
\label{apps:graph-based-regression}

\subsection{Experiment}
To investigate whether latent geometry reflects architectural design choices, we perform a graph-based analysis of embedding spaces on CIFAR-10. We consider all model embeddings associated with five macro-families introduced above: \textbf{Convolutional}, \textbf{Vision Transformer}, \textbf{Hybrid CNN--Transformer}, \textbf{Hierarchical Vision Transformer}, and \textbf{MetaFormer}. For each model, the embedding cloud induced by CIFAR-10 samples is treated as a point cloud in latent space.

To obtain a discrete approximation of the underlying representation manifold, we construct a $k$-nearest-neighbor graph with $k=10$ for each embedding cloud. This graph connects each point to its ten nearest latent neighbors and provides a local geometric skeleton from which graph-theoretic descriptors can be extracted. We then compute five graph signatures, described in Appendix~\ref{apps:graph_signatures}: \textbf{Cycle Length, Mean Square Clustering Coefficient, Wiener Index, Laplacian Eigengap}, and \textbf{Graph Diameter}. Collectively, these metrics capture complementary aspects of latent organization, including local cyclic redundancy, quadrilateral density, global compactness, spectral connectivity, and maximal geodesic extent.

Figure~\ref{fig:graph_signatures} reports the empirical distributions of these signatures across macro-families. Clear family-level differences emerge. For instance, convolutional models tend to exhibit more compact and weakly connected graphs compared to transformer-based families. Hybrid and hierarchical architectures often occupy intermediate regimes.

To quantify these differences formally, we fit a multinomial logistic regression in which the dependent variable is the model macro-family and the predictors are the graph signatures computed from each latent space, together with two scale-related covariates: \textbf{Latent Dimension} and \textbf{Number of Parameters}. This framework estimates whether systematic changes in graph geometry are associated with the probability that an embedding originates from a given architectural family, while accounting for the simultaneous contribution of all signatures.

We then assess predictor significance through variable-level likelihood-ratio tests. Concretely, for each signature, we compare the full multinomial model against a reduced model in which that single predictor is removed, while all remaining predictors are retained. If excluding a variable leads to a substantial deterioration in model likelihood, the corresponding graph signature contains information that helps discriminate among families beyond what is already explained by the others. The resulting test statistic follows a $\chi^2$ reference distribution under standard large-sample assumptions, allowing conventional $p$-value inference.


\begin{table}[t]
\centering
\begin{tabular}{lrr}
\toprule
\textbf{Variable} & \textbf{LR Statistic} & \textbf{$p$-value} \\
\midrule
Latent Dimension              & 951.78 & $1.00 \times 10^{-204}$ \\
Number of Parameters         & 455.78 & $2.44 \times 10^{-97}$ \\
Eigengap                     & 121.08 & $3.14 \times 10^{-25}$ \\
Mean Square Clustering       & 68.04  & $5.87 \times 10^{-14}$ \\
Density                      & 56.77  & $1.38 \times 10^{-11}$ \\
Cycle Length                 & 22.36  & $1.70 \times 10^{-4}$ \\
Graph Diameter               & 21.53  & $2.49 \times 10^{-4}$ \\
\bottomrule
\end{tabular}
\caption{Variable-level likelihood-ratio tests from the multinomial logistic regression predicting model macro-family from latent graph signatures and scale covariates. For each predictor, the reported statistic compares the full model against a reduced model excluding that variable only. Smaller $p$-values indicate stronger incremental explanatory value conditional on the remaining predictors.}
\label{fig:mnlogit_graph_tests}
\end{table}

Table \ref{fig:mnlogit_graph_tests} summarizes the results. Latent dimensionality and parameter count provide the strongest incremental contribution, as expected from their close relationship to model scale. More notably, several graph signatures remain highly significant after controlling for these size variables. In particular, the Laplacian Eigengap, Mean Square Clustering Coefficient, and Density exhibit strong associations with macro-family membership, indicating that architectural families differ not only in scale, but also in the connectivity and local structure of their latent manifolds. Cycle Length and Graph Diameter are also significant, though with smaller incremental contributions.

Taken together, these findings suggest that broad model families induce statistically distinguishable latent graph geometries. Convolutional and transformer-derived architectures do not merely differ in parameterization or embedding dimension; they also organize representations according to different manifold topologies, local connectivity patterns, and global geodesic structure. 

\subsection{Graph Signatures}
\label{apps:graph_signatures}

\textbf{Cycle length.} We define the average fundamental cycle length as a tree-based graph descriptor that summarizes the typical size of loops generated by redundant edges. Small values indicate mainly local cyclic structure (short loops such as triangles), whereas large values indicate longer-range cycles connecting distant regions.

Let $G=(V,E)$ be a connected graph and let $T$ be a spanning tree of $G$. For each non-tree edge $e=(u,v)\in E\setminus T$, adding $e$ to $T$ creates a unique fundamental cycle whose length is
\[
\ell_T(e)=d_T(u,v)+1,
\]
where $d_T(u,v)$ denotes the graph distance between $u$ and $v$ in $T$.

We then define
\[
\mathrm{CL}(G;T)=
\frac{1}{|E\setminus T|}
\sum_{e\in E\setminus T}\ell_T(e).
\]

Thus, this descriptor summarizes whether the graph redundancy is mostly local or global.

\textbf{Mean Square Clustering Coefficient.} The mean square clustering coefficient measures the prevalence of local four-node cycles (squares) in a graph. Let $C_4(v)$ denote the square clustering coefficient of node $v$, defined as the fraction of length-two paths through $v$ that participate in a square (see \cite{Lind_2005}). The mean square clustering coefficient is then

\[
\overline{C}_4(G)=\frac{1}{|V|}\sum_{v\in V} C_4(v).
\]

Large values indicate neighborhoods rich in local quadrilateral structure, whereas small values correspond to weak four-cycle connectivity.

\textbf{Wiener Index.} The Wiener index measures the overall distance-based compactness of a graph by summing all pairwise shortest-path distances (see \cite{Knor}). Small values correspond to compact and well-connected graphs, whereas large values indicate more elongated or weakly connected structures.

Let $d(u,v)$ denote the shortest-path distance between vertices $u,v\in V$. The Wiener index is defined as

\[
W(G)=\sum_{\{u,v\}\subseteq V} d(u,v)
\]

where only connected pairs contribute in disconnected graphs.

This descriptor provides a global summary of how efficiently vertices are mutually reachable across the graph.

\textbf{Laplacian Eigengap ($k=1$).} To provide information about global connectivity and large-scale structure of a graph, we consider spectral descriptors derived from the graph Laplacian. In particular, the Laplacian eigengap for $k=1$ (also known as the \emph{Fiedler value} \cite{fiedler1973algebraic}) measures the strength of global graph connectivity.
 
Let $L$ be the graph Laplacian matrix, and let
\[
0=\lambda_0 \leq \lambda_1 \leq \cdots \leq \lambda_{n-1}
\]
be its ordered eigenvalues. Since $\lambda_0=0$ for any graph, the eigengap at $k=1$ is

\[
\lambda_1-\lambda_0=\lambda_1.
\]

Large values indicate stronger overall connectivity, while values close to zero suggest near-disconnected components.

\textbf{Graph Diameter.} The graph diameter is a distance-based descriptor that measures the largest shortest-path separation between vertices. 
Let $d(u,v)$ denote the shortest-path distance between vertices $u,v\in V$. The graph diameter is defined as

\[
\operatorname{diam}(G)=\max_{u,v\in V} d(u,v),
\]

where only connected pairs are considered in disconnected graphs.

Small values indicate compact graphs in which all vertices are mutually close, whereas large values correspond to elongated or weakly connected structures with distant regions.


\end{document}